\documentclass[journal]{IEEEtran} 
\usepackage{amsmath,amsfonts}
\usepackage{algorithm}
\usepackage{algorithmicx}
\usepackage{algpseudocode}
\usepackage{array}
\usepackage[caption=false,font=normalsize,labelfont=sf,textfont=sf]{subfig}
\usepackage{textcomp}
\usepackage{stfloats}
\usepackage{url}
\usepackage{verbatim}
\usepackage{graphicx}
\usepackage{cite}
\usepackage{booktabs}
\usepackage{amsbsy}
\usepackage{amssymb}
\usepackage{marvosym}
\usepackage{multirow}
\usepackage{threeparttable}
\usepackage{colortbl}
\usepackage[dvipsnames]{xcolor}
\algnewcommand{\Input}{\item[\textbf{Input:}]}
\algnewcommand{\Output}{\item[\textbf{Output:}]}
\algnewcommand{\Hyperparameters}{\item[\textbf{Hyperparameters:}]}

\begin{document}

\title{RAEM: Robust Autonomous Exploration for Multi-Floor Environments with a Quadruped Robot}

\author{Zikang~Yuan$^{1,2}$, Yuan~Ren$^{3}$\textsuperscript{\Letter}, Yian~Wang$^{4}$, Yixue~Wang$^{5}$, Enze~Fang$^{5}$, Xuewei~Zhang$^{6}$, Junda~Cheng$^{5}$, \\ Chi~Chen$^{7}$, Chin-Pang~Ho$^{4}$, Lijun~Zhu$^{5}$, Shaohang~Xu$^{4}$\textsuperscript{\Letter}, Kwang-Ting Cheng$^{1,2}$ and Xin~Yang$^{5\dagger}$
	\thanks{$^{1}$AI Chip Center for Emerging Smart Systems, InnoHK Centers, Hong Kong Science Park, Hong Kong, China}%
	\thanks{$^{2}$HongKong University of Science and Technology, Hong Kong, China}%
	\thanks{$^{3}$Southeast University, Nanjing, China}%
	\thanks{$^{4}$City University of Hong Kong, Hong Kong, China}
	\thanks{$^{5}$Huazhong University of Science and Technology, Wuhan, China}%
	\thanks{$^{6}$University of Hong Kong, Hong Kong, China}%
	\thanks{$^{7}$Wuhan University, Wuhan, China}%
	\thanks{\textsuperscript{\Letter} means corresponding author.}
	\thanks{$^{\dagger}$ means project leader.}
	\thanks{E-mail: {\tt\footnotesize zikangyuan@ust.hk}, {\tt\footnotesize shaohanxu2-c@my.cityu.edu.hk}}
}

\markboth{Journal of \LaTeX\ Class Files,~Vol.~14, No.~8, August~2021}%
{Shell \MakeLowercase{\textit{et al.}}: A Sample Article Using IEEEtran.cls for IEEE Journals}


\maketitle

\begin{abstract}
In this paper, we propose RAEM, a robust autonomous exploration framework for quadruped robots operating in multi-floor environments. Most existing ground-robot exploration approaches rely on planar traversability representations, which cannot adequately represent the overlapping structures and cross-floor connectivity of multi-floor buildings. Although tomography-based representations provide effective traversability modeling for multi-floor navigation, maintaining a global tomography map incurs substantial computational overhead for online exploration with frequent replanning. Moreover, sparse and fragmented LiDAR observations in stairwells can degrade local traversability estimation, leading to irregular viewpoint placement and temporary topological disconnections. To address these challenges, RAEM adopts a hybrid local-global traversability representation, in which a local tomography map and an explicitly categorized local 3D grid map are used for online terrain analysis and connectivity evaluation, while an elevation-aware global topological graph is incrementally constructed from these local spatial representations for efficient cross-floor exploration planning. We further introduce a staircase center alignment strategy to reduce abrupt yaw variations during climbing and a dual path searching mechanism to recover guidance paths when the global topology is locally disconnected. Extensive simulation and real-world experiments demonstrate robust and computationally stable autonomous exploration across multi-floor structures, including continuous exploration of a five-floor stairwell.
\end{abstract}


\section{Introduction}
\label{Introduction}

\begin{figure}
	\begin{center}
		\includegraphics[width=\linewidth]{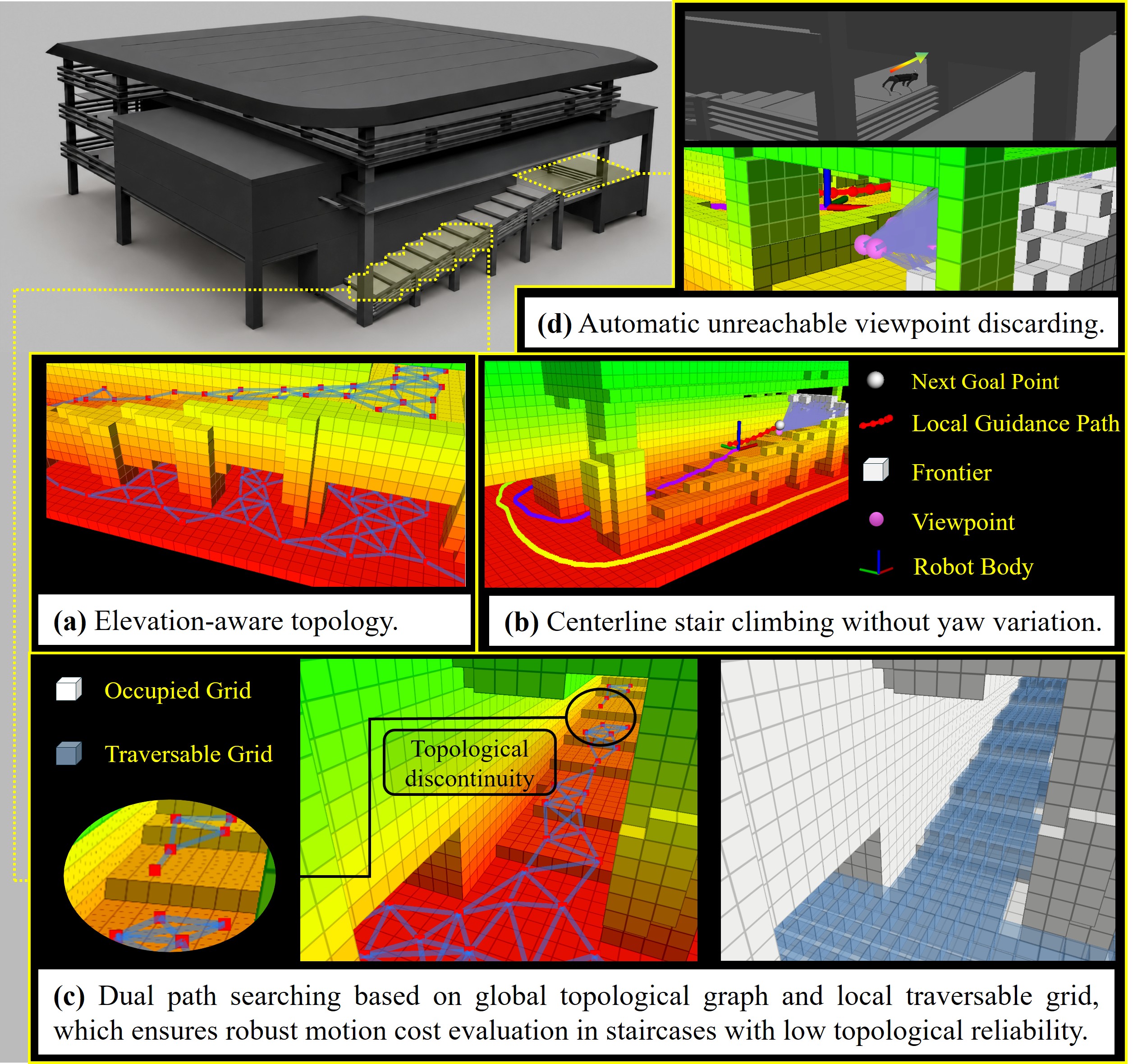}
		\caption{Illustration of the proposed autonomous exploration framework for multi-floor environments, which consists of four main capabilities: (a) online elevation-aware topological graph construction in multi-floor structures; (b) centerline-constrained viewpoint generation on staircases to prevent the quadruped robot from falling due to yaw variations on stairs; (c) traversable grid-based path searching to compensate for topological failures in stairwells due to sparse and fragmented point clouds; and (d) automatic filtering out structurally inaccessible yet physically traversable regions.}
		\label{fig1}
	\end{center}
\end{figure}

\IEEEPARstart{A}{utonomous} exploration and mapping in building environments are critical for applications such as 3D surveying \cite{yuan2022sr, yuan2024sr, zheng2024fast}, industrial inspection \cite{czimmermann2021autonomous, yang2025closed, shukla2025systematic}, and the construction of digital twins \cite{wang2025development, nochta2026participation, zhou2025multi}. Compared with aerial vehicles, ground robots offer higher payload capacities, extended endurance, and enhanced operational safety, making them particularly well-suited for indoor exploration. However, most existing exploration approaches for ground robots \cite{cao2021tare, huang2023fael, long2024hphs} rely on 2D grids \cite{patruno2021robust, ren2026real} or elevation maps \cite{fankhauser2014robot, fankhauser2018probabilistic} to assess traversability, which represent the environment using a single horizontal layer or a single-valued height field. While these simplified representations are robust and computationally efficient in single-layer planar environments, they fundamentally fail in complex multi-floor structures. As illustrated in Fig.~\ref{fig1}, such environments contain staircases connecting different floors and regions with traversable ground surfaces but remain inaccessible to the robot due to structural barriers such as railings. Conventional environmental representations restrict exploration to near-planar spaces, rendering them incapable of handling the inherent 3D complexity of multi-floor buildings.

In recent years, numerous ground robot navigation approaches \cite{frey2022locomotion, wang2023towards, yang2024efficient, li2025real, lee2025trg} have been proposed for multi-floor structural environments. Among these, PCT-Planner \cite{yang2024efficient} introduced the use of a tomography map, which consists of multiple 2.5D traversability slices, to explicitly model multi-floor traversable terrain. Despite the tomography map demonstrates significant robustness and reliability in global navigation tasks involving overhangs and multi-floor structures, this representation does not readily transfer to autonomous exploration due to two fundamental limitations. First, constructing a global tomography map and generating subsequent global guidance paths incur substantial computational overhead. In autonomous exploration, global paths must be continuously replanned in response to newly detected frontiers. This computationally expensive process conflicts with the strict timely-response requirements of exploration, rendering the global tomography representation ill-suited for autonomous exploration frameworks. Second, the quality of a tomography map is inherently contingent on the fidelity of the underlying point cloud. While global navigation assumes a complete prior map, autonomous exploration relies on incrementally built local observations. Constrained by the low mounting position of onboard LiDAR sensors, ground robots can only acquire high-fidelity observations in their immediate vicinity. This limitation is severely amplified in stairwells: when a robot approaches a staircase from below, the upward steps are observed from grazing angles, yielding sparse and fragmented point clouds. In these regions, the degraded tomography map produces irregularly scattered viewpoints and broken topological connectivity. Scattered viewpoints result in trajectories with abrupt yaw variations on staircases, and this motion pattern poses significant stability hazards for quadruped robots. Concurrently, broken connectivity prevents path searching between viewpoints on the stairs and the robot on the ground floor.

To enable robust autonomous exploration in multi-floor environments, we propose the RAEM framework. To overcome the severe computational bottlenecks associated with global tomography map, we decouple the local terrain analysis from the global traversability representation. A local tomography map and an explicitly categorized local 3D grid map are used for online terrain analysis and connectivity evaluation, while an elevation-aware global topological graph is incrementally constructed and updated for efficient cross-floor exploration planning. Specifically, the local tomography map is strictly bounded to a robot centric region, retaining only essential elements (ceiling, ground, and traversal cost) to eliminate redundant processing. From the local tomography map, we fully leverage GPU parallelization to rapidly extract a lightweight local 3D grid map that explicitly classifies spatial grids as occupied, free, or traversable. Topological vertices are preferentially sampled in regions with low traversal costs in the local tomography map. By assessing direct edge connectivity via the local 3D grid map, these extracted vertices are continuously fused into the global topological graph (as illustrated in Fig. \ref{fig1}(a)). Correspondingly, the positioning of viewpoints is also governed by these local spatial representations.

Furthermore, we propose specific strategies to address two critical issues induced by degraded tomography maps in stairwells: irregular viewpoint scattering and topological fractures. To prevent abrupt yaw variations by mitigating scattered viewpoint distributions, we introduce a staircase center alignment method. By grouping connected traversable grids at consistent heights via region growing, viewpoint positions are updated to the geometric centers of the stair treads (as illustrated in Fig. \ref{fig1}(b)). Concurrently, to resolve topological fractures (as illustrated in Fig. \ref{fig1}(c)) that cause path searching failures, the system resorts to a direct path search across the local traversable grids to generate a safe upward guidance path. As the robot climbs, the accumulating point cloud density naturally restores topological connectivity, while inherently unreachable viewpoints remain isolated and are safely discarded (as illustrated in Fig. \ref{fig1}(d)).

Extensive simulations demonstrate RAEM’s capability to explore multi-floor structures, including quadrangular spiral and corner staircases. Real-world experiments on a Unitree Go2 quadruped robot, utilizing only a standard onboard NVIDIA Jetson Orin NX 16GB processing unit and a single Mid-360S LiDAR, confirm that our method achieves complete exploration of multi-floor structures and even a five-floor stairwell, validating its robustness and practical deployability.

To summarize, the contributions of this work are fourfold:
\begin{itemize}
	\item We propose a hybrid local–global traversability representation to enable autonomous cross-floor exploration in multi-floor environments while maintaining timely online planning.
	\item We develop a highly efficient method to derive a lightweight local 3D grid map from the local tomography map. This explicitly categorized 3D grid map serves as the foundation for evaluating topological connectivity, generating viewpoints, and searching paths.
	\item We introduce targeted strategies to overcome severe perceptual degradation in stairwells. This includes a viewpoint center alignment method to prevent yaw instability during climbing, and a dual path searching mechanism that integrates global topology and local grids to ensure reliable guidance path despite topological fractures.
	\item To support community development, we commit to releasing the complete source code of RAEM upon the acceptance of this manuscript.
\end{itemize}

The rest of this paper is structured as follows. In Sec.~\ref{Related Work}, we briefly discuss the relevant literature. Sec.~\ref{Nomenclature} defines nomenclatures. Then Sec.~\ref{Methodology} details our system RAEM. Sec.~\ref{Experiments} provides experimental evaluation. Finally, we conclude the paper in Sec.~\ref{Conclusion}.

\section{Related Work}
\label{Related Work}

To contextualize the unique challenges of autonomous multi-floor exploration, this section reviews relevant literature across three intersecting domains: aerial exploration, planar ground exploration, and structural multi-floor navigation. While these fields provide fundamental insights into traversability representations and path planning, each exhibits inherent limitations when directly applied to our target problem. We review these approaches to expose the algorithmic gaps our framework resolves.

\subsection{UAV Exploration}
\label{UAV Exploration}

Unmanned aerial vehicles (UAVs) are capable of occupying any unobstructed spatial position within a 3D environment. Consequently, their traversability evaluation can rely exclusively on topological graph derived from free spaces of 3D occupancy grids, without requiring supplemental terrain representations. FUEL \cite{zhou2021fuel} is a foundational aerial framework that introduces an incrementally updated frontier information structure and a hierarchical planning strategy, utilizing Bayesian filtering over probabilistic occupancy grids to assess occupancy status. Yang et al. \cite{yang2021graph} partition the exploration map into multiple convex polyhedra, explicitly designating the geometric centers of these units as topological vertices to guide sequential exploration. To enhance occupancy grid management, Tang et al. \cite{tang2023bubble} integrate UFOMap \cite{duberg2020ufomap} alongside the concept of collision-free spheres to generate high quality topological vertices for representing traversability. Similarly, SphereMap \cite{musil2022spheremap} populates the known free space with collision-free spheres, while ECHO \cite{yu2023echo} significantly accelerates the next goal viewpoint determination process; both methods employ OctoMap \cite{hornung2013octomap} to substantially mitigate the memory consumption associated with 3D grid representation. Deviating from traditional topological graph architectures, FALCON \cite{zhang2024falcon} conducts direct A-star traversability searches on continuous 3D grids. However, this direct volumetric representation frequently induces algorithmic search timeouts in large scale scenarios. FSMP \cite{zhang2025fsmp} retains occupancy grids but introduces a mixed strategy that autonomously switches to random sampling within the free space whenever the standard path search fails. Moving away from standard occupancy grids entirely, EPIC \cite{geng2025epic} constructs an observation map directly by evaluating LiDAR view direction attributes, explicitly classifying the boundaries between high-density and low-density voxels as frontiers. Based on EPIC, Yuan et al. \cite{yuan2026aerial} enable coverage-path guidance to operate directly on point cloud maps. Despite the universal applicability of the aforementioned frameworks for UAVs, these purely volumetric representations lack the explicit terrain traversability encoding fundamentally required by ground robots.

\subsection{AGV Exploration}
\label{AGV Exploration}

Compared to aerial exploration frameworks, the fundamental distinction of autonomous ground vehicle (AGV) exploration methods lies in the explicit representation of ground traversability. However, the majority of existing approaches characterize spatial connectivity using 2D grids or elevation maps. This inherent dimensionality reduction entirely flattens multi-floor structures. TARE \cite{cao2021tare} partitions the 3D space into non overlapping subspaces and employs a hierarchical architecture where a global planner selects the optimal vertex while a local planner conducts fine grained searches. Nevertheless, it relies entirely on an elevation map, which assumes a predominantly planar scene. FAEL \cite{huang2023fael} avoids explicit environment segmentation by uniformly sampling vertices and viewpoints at fixed intervals over an elevation map. Furthermore, FAEL exhaustively conducts visibility detection between each viewpoint and frontier, which incurs large computational costs in sparsely occupied spaces. Other ground based methods similarly suffer from planar constraints. Kan et al. \cite{kan2020online} explicitly model traversability using a 2D terrain map decomposed into hexagonal cells. HPHS \cite{long2024hphs} strictly depends on a 2D grid map to evaluate structural boundaries and directly incorporates structural frontiers as topological vertices. Similarly, TIPS \cite{wang2025tips} constructs a global topological graph where vertices are incrementally generated as informative landmarks encoding historical observations. Since the traversability representations adopted by these methods are fundamentally grounded in planar scene assumptions, they are entirely precluded from application in multi-floor building structures.

\subsection{Cross-Floor Navigation}
\label{Cross Floor Navigation}

To overcome the strict planar constraints of standard ground exploration, numerous traversability representations have been developed specifically for navigation tasks involving overhanging obstacles and multi-floor structures. Miki et al. \cite{miki2022elevation} utilize slope parameters to filter out ceiling points and continuously update a local map via an overlap clearing mechanism. Meng et al. \cite{meng2023terrainnet} introduce a specific ceiling layer alongside distinct ground elevations to capture overhanging objects and varying terrain conditions. However, these augmented elevation maps remain fundamentally unsuitable for global navigation across building structures. Triebel et al. \cite{triebel2006multi} represent multi-floor scenes by recording the specific height and thickness of various surface patches into a list for each individual cell, which is an inherently complex data structure that introduces severe computational overhead during the map processing phase.  Alternatively, several methods adopt 3D occupancy grids for spatial representation. Frey et al. \cite{frey2022locomotion} employ neural networks to estimate traversability directly from occupied voxels, though this inherently provides only a coarse approximation of the underlying terrain. Wang et al. \cite{wang2023towards} utilize a valid ground filter to extract traversable regions and derive traversal costs by constructing a Euclidean Signed Distance Field. Since their trajectories are directly planned and optimized over dense voxel grids, the computational burden is significantly amplified. To unify spatial efficiency and complete 3D modeling, PCT-Planner \cite{yang2024efficient} introduces a global tomography representation based on multiple 2.5D traversability slices. While highly successful for global navigation, the continuous construction and updating of global tomography maps incur prohibitive computational overhead. Moreover, autonomous exploration inherently relies on sparse and incrementally built point clouds, particularly when observing upward staircases from grazing angles. These sparse observations completely fail to support the generation of high-quality tomography maps. Consequently, this navigation oriented traversability representation is fundamentally ill suited for direct deployment in autonomous exploration tasks.

\section{Nomenclature}
\label{Nomenclature}

\begin{table}[]
	\caption{Nomenclature}
	\label{table1}
	\centering
	\begin{tabular}{>{\centering\arraybackslash}p{1.5cm}
			|
			>{\raggedright\arraybackslash\hspace{0.5cm}}p{5.5cm}}
		\toprule
		\textbf{Symbol} & \multicolumn{1}{c}{\textbf{Description}} \\
		\hline
		$\mathcal{M}^{G}_{occ}$ & Global occupancy grid \\
		$\mathcal{M}^{L}_{tomo}$ & Local tomography map \\
		$\mathcal{M}^{L}_{grid}$ & Local 3D grid map \\
		$\mathcal{M}^{L}_{ter}$  & Inflated local 3D terrain map \\
		$\mathcal{S}$			& Discrete tomogram slices \\
		$\mathcal{T}_{obs}$		& Slice-wise obstacle trees \\
		$\mathcal{G}$           & Global topological graph \\
		$\mathcal{V}$           & Set of topological vertices \\
		$\mathcal{E}$           & Set of connective edges \\
		$\Omega_{exp}$          & Exploration task bounding box space \\
		$\Omega_{upd}$          & Dynamic map update region \\
		$\mathcal{Q}_{f}$       & Queue of newly detected frontiers \\
		$\mathcal{C}_{f}$       & Set of newly aggregated frontier clusters \\
		$\mathcal{P}_{vp}$      & Set of generated new viewpoints \\
		\bottomrule
	\end{tabular}
\end{table}

For clarity and ease of reference, we formally define the nomenclatures to be utilized in subsequent sections, all of which are represented in the world coordinate frame. The proposed framework primarily maintains two global representations. The global occupancy grid, denoted as $\mathcal{M}^{G}_{occ}$ and implemented via UFOMap \cite{duberg2020ufomap}, which employs Bayesian filtering to accumulate continuous LiDAR ray measurements. $\mathcal{M}^{G}_{occ}$ explicitly characterizes the known and unknown spaces and continuously supplies newly detected frontiers for exploration. The global topological graph is formulated as an undirected graph $\mathcal{G} = \{\mathcal{V}, \mathcal{E}\}$, where $\mathcal{V}$ represents the set of topological vertices and $\mathcal{E}$ represents the set of connective edges, collaboratively encoding the traversability across the entire environment. At the local scale, three distinct maps are dynamically constructed. The local tomography map $\mathcal{M}^{L}_{tomo}$ and the local 3D grid map $\mathcal{M}^{L}_{grid}$ are cooperatively employed to facilitate topological graph updating, viewpoint generation, and path searching prior to the execution of planning. Concurrently, the inflated local 3D terrain map $\mathcal{M}^{L}_{ter}$ is updated at a high frequency to supply geometric constraints and collision clearance information for the local planner module. $\mathcal{M}^{L}_{tomo}$ is composed of a total of $N$ discrete tomogram slices, denoted as $\mathcal{S} = \{S_k \mid k \in [0, N-1]\}$. Furthermore, to explicitly model slice-wise obstacles, grids within $\mathcal{M}^{L}_{tomo}$ possessing a traversal cost greater than a predefined barrier, are flattened into 2D coordinates and organized into spatial tree structures. This extraction yields a collection of obstacle trees denoted as $\mathcal{T}_{obs} = \{\mathcal{T}_{obs}^{k} \mid k \in [0, N-1]\}$. Geometrically, the complete 3D space bounded by the exploration task is defined as $\Omega_{exp}$, while the dynamic map update region in the immediate vicinity of the robot is denoted as $\Omega_{upd}$. Finally, regarding the exploration targets, we define the queue of newly detected frontiers as $\mathcal{Q}_{f}$, the set of newly aggregated frontier clusters as $\mathcal{C}_{f}$, and the set of generated new viewpoints as $\mathcal{P}_{vp}$. The nomenclatures and their corresponding descriptions are systematically presented in Table~\ref{table1}.

\section{Methodology}
\label{Methodology}

\begin{figure}
	\begin{center}
		\includegraphics[width=\linewidth]{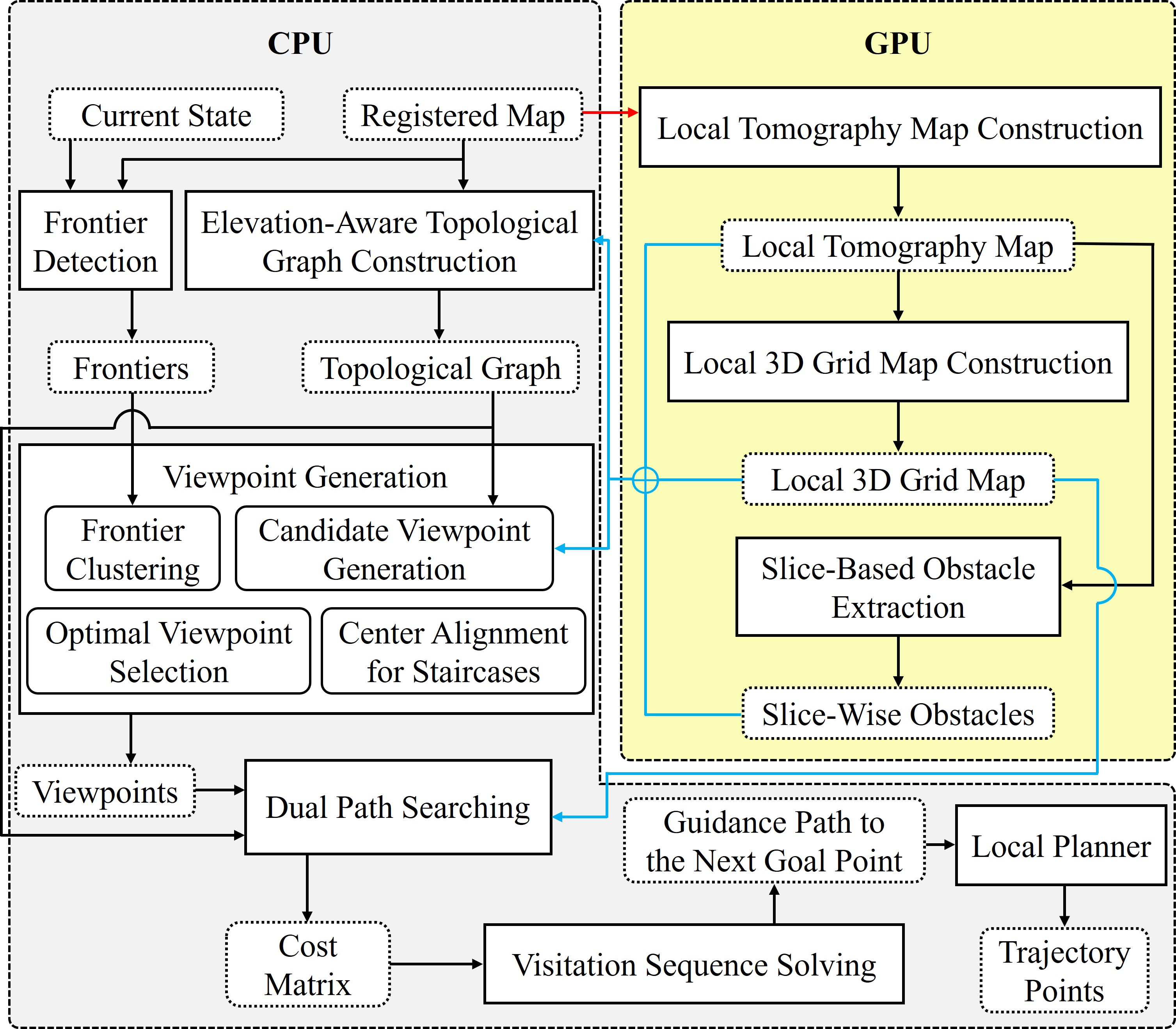}
		\caption{System overview of RAEM. The framework explicitly leverages GPU parallelization for the rapid generation of local tomography map, local 3D grid map, and the extraction of slice-based obstacles. Concurrently, the CPU manages a continuously updated occupancy grid for identifying frontiers. These detected frontiers, combined with the GPU-derived local spatial representations, directly drive the core modules, including elevation-aware topological graph construction, viewpoint generation, and dual path searching. To overcome the challenges posed by degraded tomography maps, RAEM uniquely incorporates a staircase centerline alignment method for safe viewpoint placement and a dual path searching approach to address topological fractures. The global planner subsequently determines the optimal visitation sequence of viewpoints, and the local planner generates trajectory points toward the next goal.}
		\label{fig2}
	\end{center}
\end{figure}

\subsection{System Overview}
\label{System Overview}

The system framework of RAEM is illustrated in Fig.~\ref{fig2}, which strategically distributes the computational workload across a heterogeneous computing architecture to maximize efficiency. The GPU pipeline is strictly dedicated to generating local spatial representations, ingesting the registered point cloud to efficiently construct the local tomography map $\mathcal{M}^{L}_{tomo}$. Guided by $\mathcal{M}^{L}_{tomo}$, the GPU pipeline immediately derives the explicitly categorized local 3D grid map $\mathcal{M}^{L}_{grid}$. Subsequently, it performs a slice-based obstacle extraction to construct the collection of obstacle trees $\mathcal{T}_{obs}$ for the $N$ total slices of the tomography map. These highly parallelized local spatial representations are continuously streamed to the CPU, where they underpin subsequent topology updating, viewpoint generation and path searching. Concurrently, the CPU pipeline updates the global occupancy grid $\mathcal{M}^{G}_{occ}$ to identify new frontiers $\mathcal{Q}_{f}$. By leveraging the GPU-supplied $\mathcal{M}^{L}_{tomo}$ and $\mathcal{M}^{L}_{grid}$, the CPU pipeline executes elevation-aware topological graph construction to update the global topological graph $\mathcal{G}$. Within the viewpoint generation module, the newly detected frontiers $\mathcal{Q}_{f}$ are grouped into frontier clusters $\mathcal{C}_{f}$ based on their spatial adjacency in Euclidean space. For each frontier cluster, the optimal observation viewpoint is selected from the candidate set, whose position is strictly constrained by a specialized staircase center alignment method to guarantee motion safety in stairwells. The dual path searching approach first searches path from the current robot position to each viewpoint within $\mathcal{P}_{vp}$, along with the pairwise paths among all viewpoints, to evaluate motion costs. Utilizing the comprehensive cost matrix, the global planner subsequently solves the optimal visitation sequence to determine the final exploration targets and feeds a guidance path to the local planner, which relies on the inflated local 3D terrain map $\mathcal{M}^{L}_{ter}$ to generate executable trajectory points for the quadruped robot.

\subsection{Frontier Detection}
\label{Frontier Detection}

We configure the perception range as a spherical region with a predefined radius $R_{occ}$ centered at the LiDAR position. As the spatial occupancy states are periodically updated, the bounding box enclosing all the modified grids is designated as $\Omega_{upd}$. Subsequently, we extract new frontiers for the grids situated exactly at the boundary separating the free and unknown spaces from $\Omega_{upd}$, and append them to $\mathcal{Q}_{f}$. Since the fundamental operations regarding this specific procedure are implemented comprehensively in Tang et al. \cite{tang2023bubble} and FAEL \cite{huang2023fael}, we omit the details here for brevity.

\subsection{Local Tomography Map Construction}
\label{Local Tomography Map Construction}

\begin{figure}
	\begin{center}
		\includegraphics[width=\linewidth]{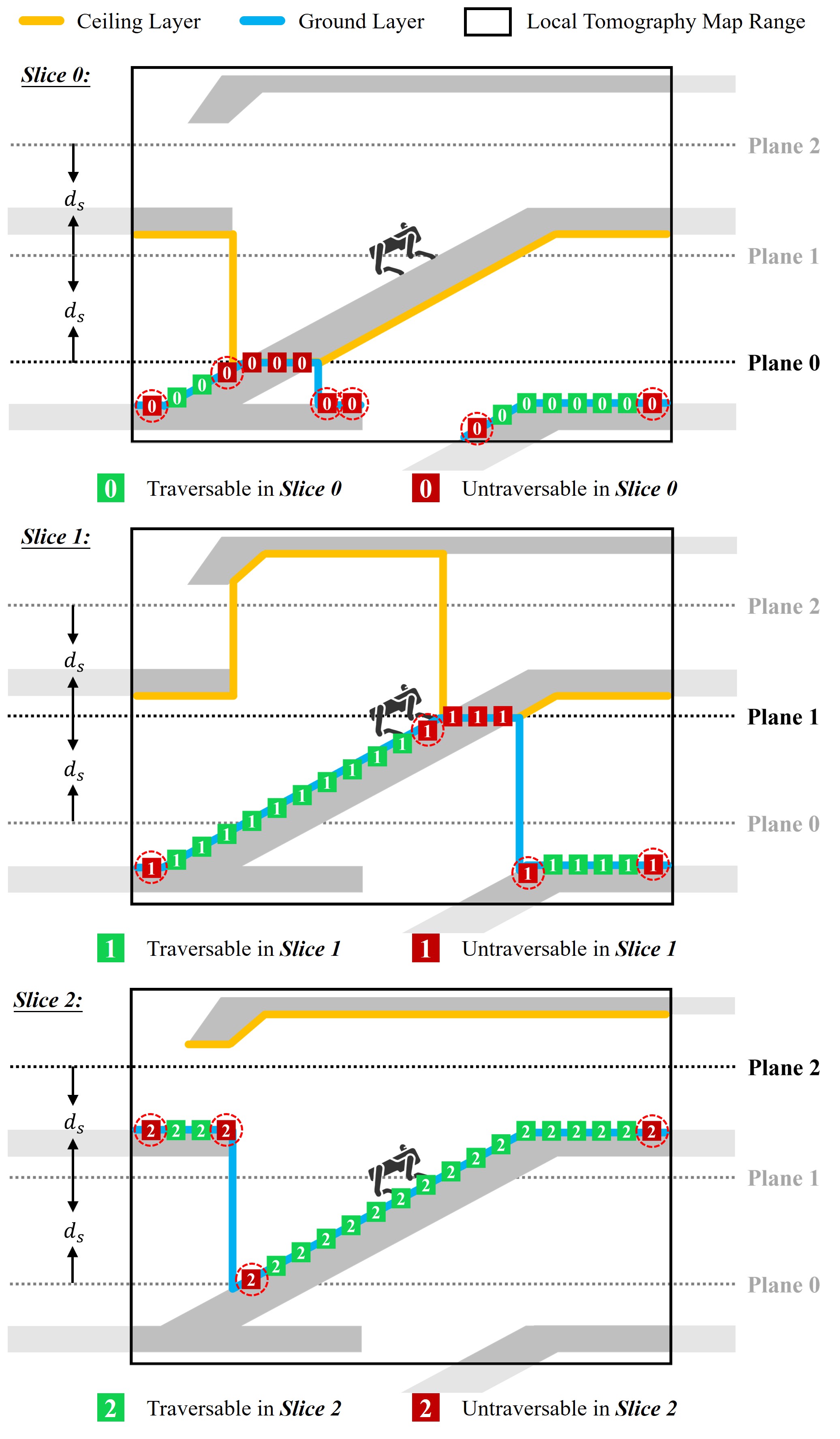}
		\caption{Traversability analysis of local tomogram slices. Slices are built by projecting the local point cloud onto equidistant horizontal planes, capturing orange ceiling and blue ground layers. The squares represent the cell traversability (green:
			traversable, red: untraversable) considering the ground conditions and the ceiling height. The red dotted circles highlight cells conservatively marked as untraversable due to strict proximity to structural edges or point cloud map boundaries, despite being physically traversable. Beneath staircases, sparse point cloud observations combined with this conservative labeling generate massive falsely untraversable cells, severely compromising topological building, viewpoint generation, and path searching.}
		\label{fig3}
	\end{center}
\end{figure}

Following the tomography map construction introduced in PCT-Planner \cite{yang2024efficient}, we construct a simplified local variant, denoted by $\mathcal{M}^{L}_{tomo}$, for online terrain evaluation. We evaluate a fixed bounding box $\mathcal{B}_{local}$ centered on the robot and construct the tomography map entirely from scratch at each cycle, which fully leverages the parallelization capabilities of GPU. Additionally, we compress the information density of the map by restricting each tomogram slice to record only ceiling, ground, and traversal cost. This highly simplified configuration effectively minimizes both the overall computational expenditure required for tomography map construction and the data transfer costs between different modules.

Formally, the local tomography map $\mathcal{M}^{L}_{tomo}$ is composed of a total of $N$ discrete 2.5D traversability slices (as illustrated in Fig.~\ref{fig3}), denoted as $\{S_k \mid k \in [0, N-1]\}$, which are distributed equidistantly with a vertical interval $d_s = [\mathcal{B}_{local}]_z / N$ above and below the horizontal plane centered at the current robot position. Each slice $S_k$ is structured as a multi-channel cell map with a spatial resolution of $r_{tomo}$. Mathematically, a single slice is defined as $S_k=\{\boldsymbol{e}_k^G, \boldsymbol{e}_k^C, \boldsymbol{c}_k^T\}$, which encompasses a ground elevation layer $\boldsymbol{e}_k^G=\{e_{i, j, k}^G\}$ and a ceiling elevation layer $\boldsymbol{e}_k^C=\{e_{i, j, k}^C\}$ to explicitly encode the terrain conditions and overhanging structures, respectively. Additionally, it incorporates a corresponding traversal cost map $\boldsymbol{c}_k^T = \{c_{i, j, k}^T\}$. $e$ and $c$ respectively denote the absolute elevation and the traversal cost values of an individual cell, while $i$, $j$, and $k$ correspond to the row, column, and slice indices.

To construct each slice $S_k$, its corresponding horizontal plane partitions the local point cloud into a lower group and an upper group, which respectively contain the points located strictly below and above the plane. For the generation of the ground elevation layer $\boldsymbol{e}_k^G$, all points within the lower group are orthogonally projected upwards onto this plane. The projection depth of each point is calculated as its absolute vertical distance to the plane. Following a spatial discretization process, the ground elevation value $e_{i,j,k}^G$ for each cell is computed by subtracting the minimum projection depth among all its enclosed points from the absolute height of the plane. The ceiling elevation layer $\boldsymbol{e}_k^C$ is constructed symmetrically. Points in the upper group are orthogonally projected downwards, and the cell elevation $e_{i,j,k}^C$ is obtained by adding the minimum projection depth to the plane height. Cells devoid of any projected points are explicitly assigned an invalid value for their corresponding $e_{i,j,k}^G$ or $e_{i,j,k}^C$. By executing this systematic procedure, every horizontal plane successfully generates a complementary pair of ground and ceiling layers for its associated tomogram slice.

\begin{figure*}
	\begin{center}
		\includegraphics[width=\linewidth]{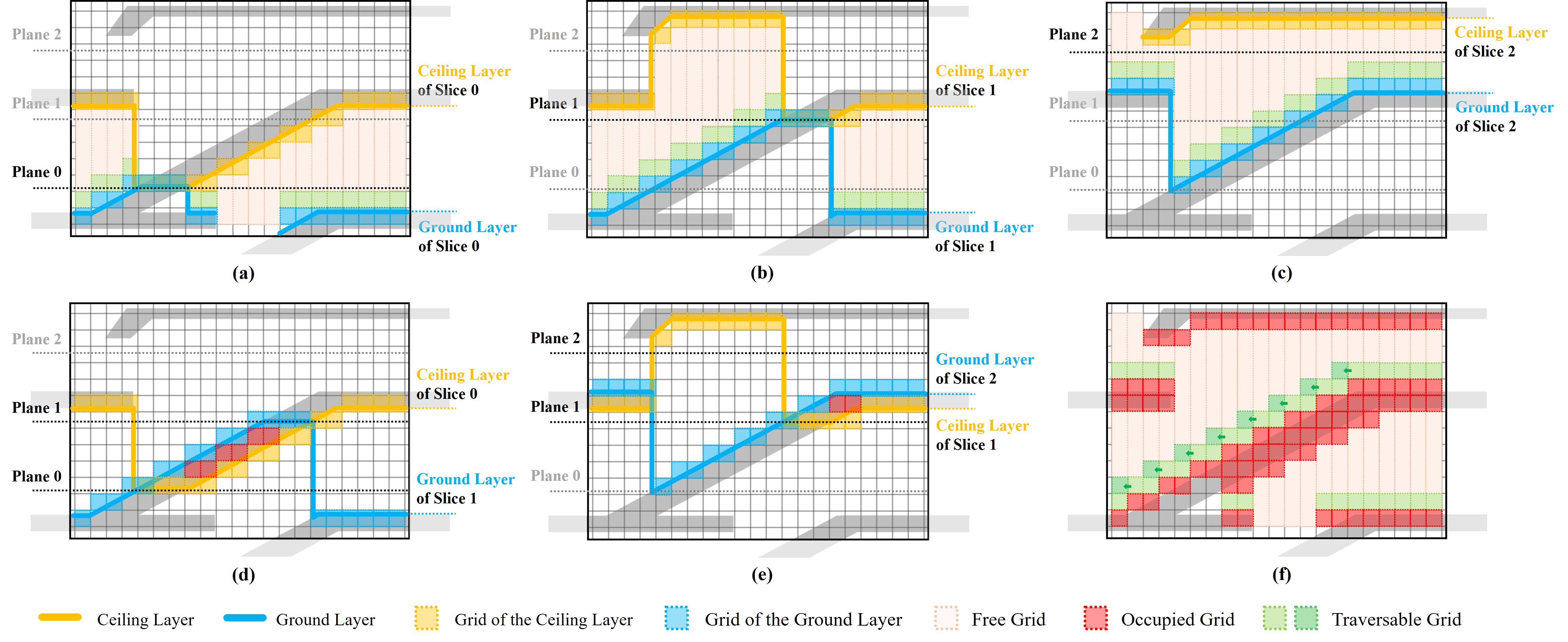}
		\caption{Illustration of construction processes of the local 3D grid map $\mathcal{M}^{L}_{grid}$. (a)-(c) illustrate the intermediate grid state assignments processed through the structural logic of individual slices. (d) and (e) demonstrate the structural logic across multiple slices, where previously unassigned spatial grids bounded between the ceiling of a lower slice and the ground of an adjacent upper slice are rigorously classified as occupied grids (red) representing solid architectural structures. (f) displays the final integrated map, highlighting a morphological horizontal dilation applied to the traversable grids (indicated by dark green arrows) to significantly reinforce the robustness of path searching for staircases.}
		\label{fig4}
	\end{center}
\end{figure*}

To comprehensively evaluate the traversability of each cell, the vertical spatial clearance and the geometric ground roughness are concurrently analyzed. By evaluating the vertical difference between each corresponding pair of ground and ceiling layers, we determine the clearance interval $\boldsymbol{d}^I = \boldsymbol{e}^C - \boldsymbol{e}^G$. Simultaneously, for the ground elevation $\boldsymbol{e}^G$, we utilize the finite difference method along the orthogonal $x$ and $y$ axes to obtain the maximum directional gradient magnitude $m^{xy} = \max(\vert{}g^x\vert{}, \vert{}g^y\vert{})$ and the overall gradient magnitude $m^{grad} = \sqrt{(g^x)^2 + (g^y)^2}$. Furthermore, to evaluate if steep geometric edges are safe to step on, we calculate the percentage $p_s$ of neighboring grids possessing $m^{grad} < \theta_s$ within a localized spatial patch.

Ultimately, the comprehensive traversal cost $c_{i, j, k}^T$ integrates all physical constraints and energy consumption penalties into a single unified formulation. A cell is deemed traversable only if its vertical clearance satisfies the minimum physical limit $d_{min}$ and its geometric roughness meets specific safety criteria. For these safely traversable regions, the algorithm dynamically imposes a penalty by selecting the maximum cost between posture deformation and terrain roughness. We formally define the necessary posture deformation as the height adjustment magnitude $\Delta d = d_{ref} - d^I$ ($d^I \in \boldsymbol{d}^I$). Lowering the robot body height intrinsically increases energy consumption, which is linearly penalized by scaling this deformation $\Delta d$ with a proportional factor $\alpha_d$. Concurrently, navigating rough terrain incurs geometric penalties scaled by $\alpha_s$ for gentle surfaces and $\alpha_b$ for traversable edges. If a cell fails to meet these specific safety conditions, it is strictly penalized by the barrier cost $c^B$. Consequently, the unified traversal cost is formulated as:
\begin{equation}
\label{equation1}
\small
	c_{i, j, k}^T = \begin{cases} \max \left( \alpha_d \Delta d, \, \alpha_s \left(\frac{m^{grad}}{\theta_s}\right)^2 \right) & \text{if } d^I \ge d_{min} \\  & \land m^{grad} < \theta_s \\ \max \left( \alpha_d \Delta d, \, \alpha_b \left(\frac{m^{xy}}{\theta_b}\right)^2 \right) & \text{if } d^I \ge d_{min} \\  & \land m^{xy} \le \theta_b \land p_s > \theta_p \\ c^B & \text{otherwise} \end{cases}
\end{equation}
where $d_{min}$ represents the minimum physical limit required for the vertical clearance, $\theta_s$ denotes the gradient threshold for identifying traversable gentle surfaces, $\theta_b$ denotes the structural barrier threshold for the directional gradient, and $\theta_p$ defines the minimum required percentage of safe stepping neighbors to guarantee a traversable edge region. As illustrated by the red dotted circles in Fig.~\ref{fig3}, the traversable cells computed according to Eq.~\ref{equation1} are highly conservative. Consequently, cells within the ground layer located in close proximity to structural edges are conservatively marked as untraversable to guarantee safety. Upon the complete construction of $\mathcal{M}^{L}_{tomo}$, we define a boolean indicator function to efficiently query the traversability status of any specific cell. Mathematically, this querying mechanism is formulated as:
\begin{equation}
\label{equation2}
\small
	\text{isCellTraversable}(i, j, k) =  \begin{cases}  \text{true} & \text{if } c_{i, j, k}^T < c^B \\ \text{false} & \text{otherwise} \end{cases}
\end{equation}

\subsection{Local 3D Grid Map Construction}
\label{Local 3D Grid Map Construction}

The 2.5D slice representation of the tomography map cannot be directly utilized to evaluate the spatial connectivity between topological vertices, nor between topological vertices and generated viewpoints. Furthermore, it is fundamentally inadequate for path searching when topological fractures occur. Consequently, we derive an explicitly categorized local 3D grid map $\mathcal{M}^{L}_{grid}$ from the multiple 2.5D traversability slices. Recognizing the massive computational overhead associated with dense 3D grid updates, we implement a highly parallelized GPU architecture. The construction algorithm assigns an individual GPU thread to process each vertical spatial column sequentially, querying the corresponding multiple data channels from the tomogram slices $\mathcal{S} = \{S_k \mid k \in [0, N-1]\}$.

Each grid $w_{i,j,l}$ within $\mathcal{M}^{L}_{grid}$ maintains a discrete spatial state defined as $\sigma_{i,j,l} \in \{\sigma_{unk}, \sigma_{occ}, \sigma_{free}, \sigma_{trav}\}$, representing unknown, occupied, free, and traversable spaces respectively. All grids are systematically initialized to the default state $\sigma_{unk}$. For a specific vertical column located at the planar indices $(i,j)$, the assigned GPU thread evaluates each vertical grid $w_{i,j,l}$ in an ascending sequential order, strictly commencing from the lowest grid where $l=0$. The absolute geometric elevation $z_l$ of the current target grid $w_{i,j,l}$ is calculated as:
\begin{equation}
\label{equation3}
\small
	z_l = z_{min} + l \times r_{grid} \quad (r_{grid} = r_{tomo})
\end{equation}
where $z_{min}$ is the vertical map boundary minimum, and $r_{grid}$ is the volumetric spatial resolution of $\mathcal{M}^{L}_{grid}$.

\begin{figure*}
	\begin{center}
		\includegraphics[width=\linewidth]{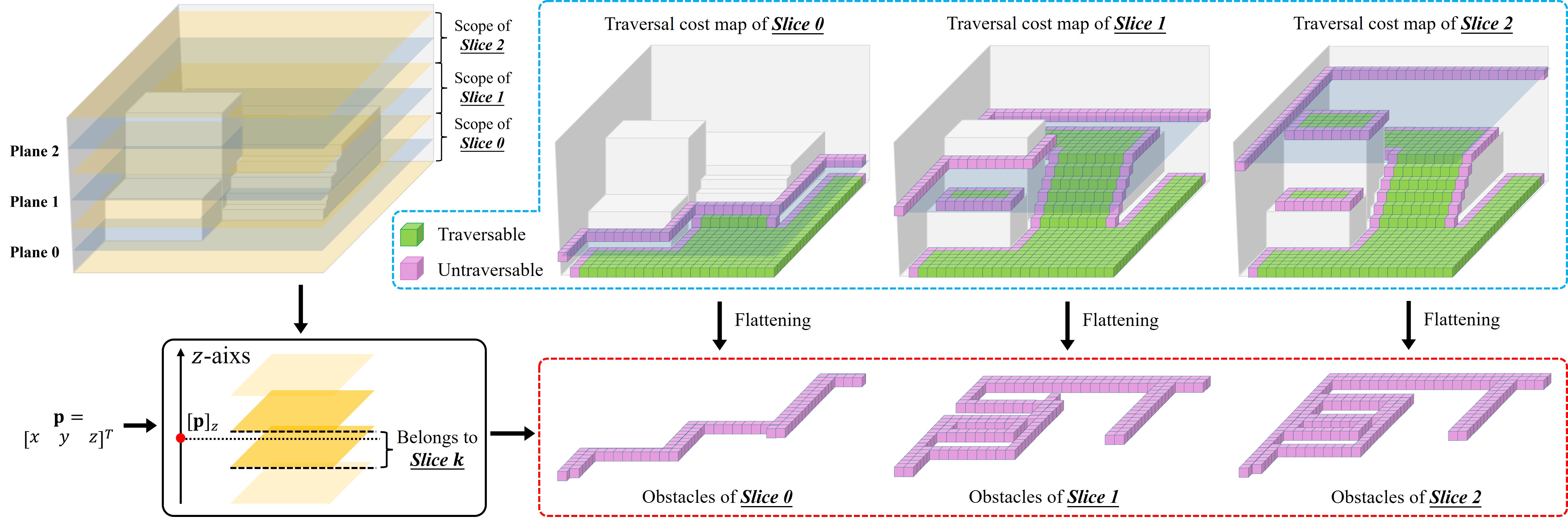}
		\caption{Overview of the slice-based obstacle extraction and distance-to-obstacle query. The multi channel tomogram slices explicitly encode traversable regions (green cubes) and untraversable obstacle boundaries (purple cubes). The untraversable cells within each slice $S_k$ undergo a dimensional flattening operation, directly projecting the complex spatial structures into a dense 2D point cloud to construct a dedicated planar tree $\mathcal{T}_{obs}^k$. For any query point $\mathbf{p}$, its corresponding slice index $k$ is mathematically determined based solely on its vertical elevation component $[\mathbf{p}]_z$, subsequently enabling rapid and highly efficient distance-to-obstacle evaluations through $\mathcal{T}_{obs}^k$.}
		\label{fig5}
	\end{center}
\end{figure*}

We first handle the structural logic inherent to a single slice. All available slices are traversed until the target elevation $z_l$ satisfies the condition $z_l \ge e^G_{i,j,k}$ and $z_l \le e^C_{i,j,k}$. If all available slices fail to fulfill this geometric condition, the grid $w_{i,j,l}$ is incorporated into the unassigned set $\mathcal{W}_{void}$, entirely bypassing the structural logic of a single slice. Conversely, once a valid slice $S_k$ satisfying these criteria is successfully identified, we calculate the discrete vertical grid indices for both the ground elevation $e^G_{i,j,k}$ and the ceiling elevation $e^C_{i,j,k}$ within the local 3D grid map $\mathcal{M}^{L}_{grid}$:
\begin{equation}
\label{equation4}
\small
	\begin{aligned}
		l^G &= \lfloor (e^G_{i,j,k} - z_{min}) / r_{grid} \rfloor \\
		l^C &= \lfloor (e^C_{i,j,k} - z_{min}) / r_{grid} \rfloor
	\end{aligned}
\end{equation}
The spatial state $\sigma_{i,j,l}$ is then accordingly updated based on the vertical index of $w_{i,j,l}$:
\begin{equation}
\label{equation5}
\small
	\sigma_{i,j,l} =
	\begin{cases}
		\sigma_{occ}  & \text{if } l = l^G \text{ or } l = l^C \\
		\sigma_{trav} & \text{if } l = l^G + 1 \\
		\sigma_{free} & \text{otherwise}
	\end{cases}
\end{equation}
If the target grid $w_{i,j,l}$ exactly shares the identical spatial grid index with either the ground or the ceiling surface, the state is rigidly marked as $\sigma_{occ}$. Conversely, if the grid elevation resides strictly in the intermediate space bounded between the ground and ceiling structures, a traversability evaluation is initiated: If the grid is situated exactly one grid unit above the ground layer, mathematically satisfying $l = l^G + 1$, the state is updated to $\sigma_{trav}$. This explicitly signifies that the grid is completely unobstructed and serves as a valid stepping location reachable by the quadruped robot. Otherwise, for any higher grids satisfying $l > l^G + 1$, the state is assigned as $\sigma_{free}$. This indicates that while the volumetric space is safely unobstructed, its elevated distance from the ground renders it physically inaccessible for quadruped locomotion. Fig.~\ref{fig4} (a)-(c) illustrate the intermediate mapping result for each slice after applying the structural logic of a single slice. It is evident that while the majority of the grids are successfully assigned accurate spatial states, a specific fraction remains unclassified.

To comprehensively characterize the states of all grids within $\mathcal{B}_{local}$, we execute further evaluations on the elements housed in $\mathcal{W}_{void}$ according to the structural logic across multiple slices. For each grid $w_{i,j,l}$ belonging to the unassigned set $\mathcal{W}_{void}$, the algorithm traverses all available slices until the target elevation $z_l$ satisfies the condition $z_l > e^C_{i,j,k}$ and $z_l < e^G_{i,j,k+1}$. As illustrated in Fig.~\ref{fig4} (d) and (e), the grids fulfilling this geometric condition are situated in the intermediate space bounded by the ceiling of the lower slice $k$ and the ground of the adjacent upper slice $k+1$. These specific locations typically represent the solid structural regions separating different floors. Consequently, the spatial state $\sigma_{i,j,l}$ for these grids is explicitly updated to $\sigma_{occ}$.

Fig.~\ref{fig4} (f) illustrates the final constructed local 3D grid map $\mathcal{M}^{L}_{grid}$ after integrating the intermediate outcomes presented in Fig.~\ref{fig4} (a)-(e). To enhance the robustness of the subsequent path searching based on $\mathcal{M}^{L}_{grid}$, we apply a single grid horizontal dilation to the traversable grids, denoted by the dark green arrows in Fig.~\ref{fig4} (f). Crucially, this horizontal expansion is executed only if the vertically adjacent grid immediately beneath the dilated location possesses a spatial state of either $\sigma_{trav}$ or $\sigma_{occ}$. This morphological operation significantly reinforces the traversability representation, particularly within geometrically complex staircase regions. It is highly important to emphasize that the entire construction of $\mathcal{M}^{L}_{grid}$ relies exclusively on the geometric elevation data extracted from the ceiling layers $\boldsymbol{e}_k^C$ and ground layers $\boldsymbol{e}_k^G$ of $\mathcal{M}^{L}_{tomo}$. By deliberately omitting the traversal cost layers $\boldsymbol{c}_k^T$ calculated under the strictly conservative safety criteria, the traversable spatial grids identified within $\mathcal{M}^{L}_{grid}$ represent a more permissive traversable space compared to the highly restricted traversable cells in $\mathcal{M}^{L}_{tomo}$.

Upon the complete construction of $\mathcal{M}^{L}_{grid}$, we first formulate a spatial mapping function to convert any continuous Euclidean coordinate $\mathbf{p} = [x, y, z]^T$ into its corresponding discrete grid indices:
\begin{equation}
\label{equation6}
\small
	\mathcal{I}(\mathbf{p}) = \left( \lfloor \frac{x - x_{min}}{r_{grid}} \rfloor, \lfloor \frac{y - y_{min}}{r_{grid}} \rfloor, \lfloor \frac{z - z_{min}}{r_{grid}} \rfloor \right)
\end{equation}
where $x_{min}, y_{min}$, and $z_{min}$ are the boundary minimum coordinates of $\mathcal{M}^{L}_{grid}$. Building upon this mapping operator, we define a state retrieval function to precisely query the exact status of any Euclidean location within $\mathcal{B}_{local}$. By explicitly applying the discretization operator to the input coordinate, the function is rigorously formulated as:
\begin{equation}
\label{equation7}
\small
	\text{getGridState}(\mathbf{p}) = \sigma_{\mathcal{I}(\mathbf{p})}
\end{equation}
Furthermore, we introduce a boolean function to determine whether the direct path is completely unobstructed, mathematically formulated as:
\begin{equation}
\label{equation8}
\small
	\begin{split}
		\text{isCollisionFree}(\mathbf{p}_1, \mathbf{p}_2) = 
		\begin{cases} 
			\text{true} & \forall \mathbf{p} \in \overline{\mathbf{p}_1 \mathbf{p}_2}, \\[2pt]
			& \text{getGridState}(\mathbf{p}) \\[2pt] & \in \{\sigma_{free}, \sigma_{trav}\} \\[2pt]
			\text{false} & \text{otherwise}
		\end{cases}
	\end{split}
\end{equation}
where $\overline{\mathbf{p}_1 \mathbf{p}_2}$ explicitly represents the set of all continuous coordinates along the linear segment connecting the query points $\mathbf{p}_1$ and $\mathbf{p}_2$. This formal definition strictly guarantees that the function returns true if and only if every individual coordinate evaluated along the connecting trajectory maps exclusively to a spatial state of either $\sigma_{free}$ or $\sigma_{trav}$. If any intersected grid resolves to $\sigma_{occ}$ or $\sigma_{unk}$, the geometric condition fails and the function immediately returns false.

\subsection{Slice-Based Obstacle Extraction}
\label{Slice-Based Obstacle Extraction}

To efficiently support distance-to-obstacle queries across varying elevations, we explicitly extract the obstacle boundaries from the traversal cost map of each individual slice. As illustrated in Fig.~\ref{fig5}, the multi-channel tomogram slices inherently encode both the traversable cells represented by the green cubes and the untraversable cells represented by the purple cubes. These untraversable cells undergo a dimensional flattening operation. Specifically, for an untraversable cell located at discrete matrix indices $(i, j)$ within a given slice $S_k$, its continuous spatial coordinate $\mathbf{p}_{flat} = [p_x, p_y]^T$ is mathematically projected utilizing the map resolution $r_{grid}$ and the local map center $(x_c, y_c)$ along with the total grid dimensions $D_x$ and $D_y$:
\begin{equation}
\label{equation9}
\small
	\begin{split}
		p_x &= x_c + (i - \frac{D_x}{2}) r_{grid} \\
		p_y &= y_c + (j - \frac{D_y}{2}) r_{grid}
	\end{split}
\end{equation}
This flattening process directly reduces the complex spatial boundaries of each slice into a dense 2D point cloud. Subsequently, the flattened obstacle point cloud for $S_k$ is seamlessly integrated into a planar tree $\mathcal{T}_{obs}^k$. This hierarchical tree structure enables rapid distance queries from any 3D point to the nearest obstacle boundary across different slices. Specifically, when evaluating the spatial safety of a query point $\mathbf{p} = [x, y, z]^T$, the algorithm first determines its corresponding slice index $k$ relying strictly on its vertical elevation component $[\mathbf{p}]_z$:
\begin{equation}
\label{equation10}
\small
	k = \lfloor \frac{[\mathbf{p}]_z - z_{min}}{d_s} \rfloor
\end{equation}
where $z_{min}$ represents the vertical map boundary minimum and $d_s$ represents the vertical interval separating adjacent slices. Following the slice index identification, the system executes fast nearest neighbor distance queries against the specific planar tree $\mathcal{T}_{obs}^k$. We define a dedicated distance evaluation function that accepts the selected obstacle tree and the planar components $[\mathbf{p}]_{x,y}$ of the query point as inputs:
\begin{equation}
\label{equation11}
\small
	d_{obs} = \text{getObstacleDistance}(\mathcal{T}_{obs}^k, [\mathbf{p}]_{x,y})
\end{equation}
Eq.~\ref{equation11} directly returns the minimum distance to obstacles, thereby ensuring rigorously safe topological vertex generation and candidate viewpoint selection.

\subsection{Elevation-Aware Topological Graph Construction}
\label{Elevation-Aware Topological Graph Construction}

\begin{figure}
	\begin{center}
		\includegraphics[width=\linewidth]{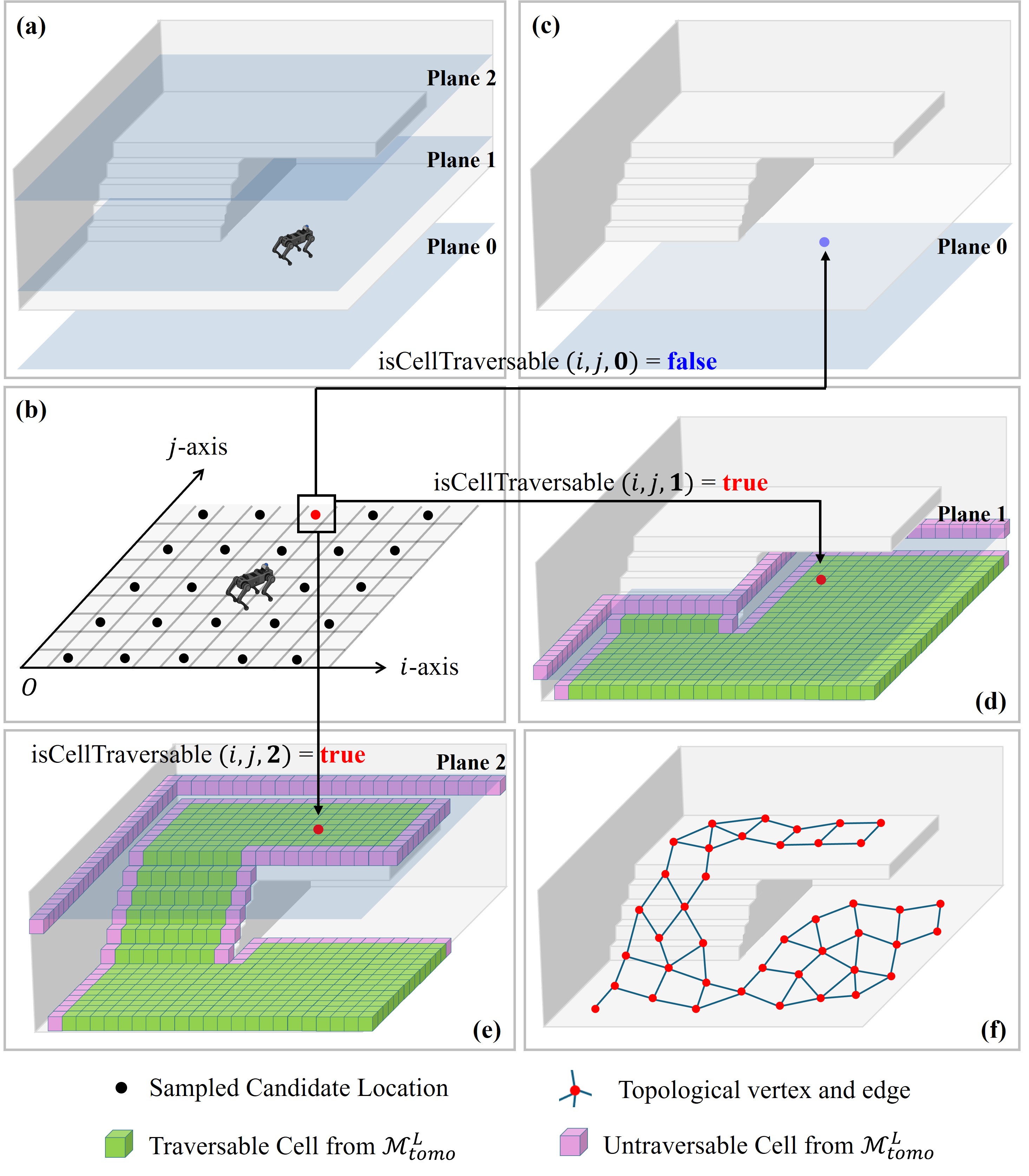}
		\caption{Overview of the elevation-aware topological graph construction. (a) A spatial environment featuring overlapping elevation layers. (b) Uniformly sampling candidate locations across the 2D grid plane $(i, j)$. (c) Untraversable or unavailable cells on Plane 0 generate no vertices. (d) and (e) Structurally safe and traversable cells on Plane 1 and Plane 2 successfully instantiate topological vertices at their respective ground elevations. (f) The finalized topological graph interconnecting these generated vertices utilizing bidirectional collision-free spatial edges under maximum allowable edge degree limitations.}
		\label{fig6}
	\end{center}
\end{figure}

\begin{algorithm}[t]
	\caption{Elevation-Aware Topological Graph Construction}
	\label{alg:topology}
	\small
	\begin{algorithmic}[1]
		\Input Local tomography map $\mathcal{M}^{L}_{tomo}$, local 3D grid map $\mathcal{M}^{L}_{grid}$, current global topology graph $\mathcal{G} = (\mathcal{V}, \mathcal{E})$, slice-wise obstacle trees $\mathcal{T}_{obs} = \{\mathcal{T}_{obs}^k\}$, current robot position $\mathbf{p}_{curr}$
		\Output Updated global topology graph $\mathcal{G}$
		\Hyperparameters Maximum sampling radius $R_{topo}$, minimum vertex spacing $r_{min}$, maximum allowable edge degree $D_{max}$, safety margin to obstacles $d_{safe}$
		\State $\mathcal{V}_{new} \leftarrow \emptyset$ 
		\State $\mathcal{V}_{local} \leftarrow \{v \in \mathcal{V} \mid \Vert{}v - \mathbf{p}_{curr}\Vert{} \le R_{topo}\}$
		\State 
		\State \textbf{Phase 1: Vertex Generation}
		\For{each sampled horizontal index $(i, j)$ around $\mathbf{p}_{curr}$}
		\For{each tomogram slice index $k \in [0, N - 1]$}
		\If{$\text{isCellTraversable}(i, j, k)$ is true}
		\State $[p_x, p_y]^T \leftarrow \mathcal{I}^{-1}(i, j)$
		\State Generate candidate vertex $\mathbf{p} \leftarrow [p_x, p_y, e^G_{i,j,k}]^T$
		\State $d_{obs} \leftarrow \text{getObstacleDistance}(\mathcal{T}_{obs}^k, [\mathbf{p}]_{x,y})$
		\If{$d_{obs} > d_{safe}$}
		\If{$\forall v \in (\mathcal{V}_{new} \cup \mathcal{V}_{local}), \, \Vert{}\mathbf{p} - v\Vert{} > r_{min}$}
		\State $\mathcal{V}_{new} \leftarrow \mathcal{V}_{new} \cup \{\mathbf{p}\}$
		\State Add new vertex $\mathbf{p}$ to $\mathcal{V}$
		\EndIf
		\EndIf
		\EndIf
		\EndFor
		\EndFor
		\State 
		\State \textbf{Phase 2: Edge Connection}
		\State $\mathcal{V}_{com} \leftarrow \mathcal{V}_{new} \cup \mathcal{V}_{local}$
		\State Construct $\mathcal{T}_{local}$ containing the comprehensive set $\mathcal{V}_{com}$
		\For{each specific vertex $v \in \mathcal{V}_{com}$}
		\State Query all near vertices $\mathcal{N}_v$ via rapid neighborhood search utilizing $\mathcal{T}_{local}$
		\For{each neighbor candidate $n \in \mathcal{N}_v$}
		\If{$\text{degree}(v) \ge D_{max}$}
		\State \textbf{break}
		\EndIf
		\If{$\text{degree}(n) < D_{max}$ \textbf{and} edge $(v, n) \notin \mathcal{E}$}
		\If{$\text{isCollisionFree}(v, n)$ is true}
		\State Add bidirectional topological edge $(v, n)$ to $\mathcal{E}$
		\State Update $\text{degree}(v)$ and $\text{degree}(n)$
		\EndIf
		\EndIf
		\EndFor
		\EndFor
		\State 
		\Return $\mathcal{G}$
	\end{algorithmic}
\end{algorithm}

Based on the local tomography map $\mathcal{M}^{L}_{tomo}$, the local 3D grid map $\mathcal{M}^{L}_{grid}$ and the slice-wise obstacle trees $\mathcal{T}_{obs}$, we incrementally construct an elevation-aware global topological graph $\mathcal{G} = (\mathcal{V}, \mathcal{E})$ to represent traversable connectivity across different elevation layers. Prior to the vertex generation, we extract a localized subset of existing vertices $\mathcal{V}_{local}$ from the global graph $\mathcal{G}$. This extraction is bounded by a spherical perception radius $R_{topo}$ centered at the current robot position $\mathbf{p}_{curr}$:
\begin{equation}
	\label{equation12}
	\small
	\mathcal{V}_{local} = \{v \in \mathcal{V} \mid \Vert{}v - \mathbf{p}_{curr}\Vert{} \le R_{topo}\}
\end{equation}
Following this initialization, the vertex generation process systematically evaluates the spatial geometry across all discrete elevation layers. As illustrated in Fig.~\ref{fig6} (a) and (b), the algorithm initiates by uniformly sampling horizontal locations around $\mathbf{p}_{curr}$ across the 2D grid plane, corresponding to the discrete spatial indices $(i, j)$. For every sampled horizontal location, the algorithm iterates vertically through all available tomogram slices along the $k$ dimension. At each specific spatial index $(i, j, k)$, the structural status is rigorously assessed utilizing Eq.~\ref{equation2}. If the cell is classified as traversable, a candidate coordinate $\mathbf{p} = [p_x, p_y, e^G_{i,j,k}]^T$ is initially generated, where $p_x$ and $p_y$ denote the Euclidean coordinates obtained via the inverse mapping of the first two dimensions from Eq.~\ref{equation6}. Subsequently, to guarantee rigorous spatial safety for the quadruped platform, this candidate $\mathbf{p}$ must undergo a secondary clearance verification. We explicitly evaluate its minimum distance to the nearest obstacle boundary utilizing Eq.~\ref{equation11}. The candidate is formally accepted only if this evaluated distance exceeds a predefined safety margin $d_{safe}$. To prevent excessive vertex density and redundant computations, this validated candidate $\mathbf{p}$ is formally instantiated and appended to the newly generated vertex set $\mathcal{V}_{new}$ if its Euclidean distance to all existing vertices in the combined set $\mathcal{V}_{new} \cup \mathcal{V}_{local}$ exceeds a minimum spatial interval $r_{min}$. Conversely, if the cell is classified as untraversable, unavailable, or fails the obstacle clearance check, as depicted for Plane 0 in Fig.~\ref{fig6} (c), the location is deemed unsafe and no vertex is created. This vertical iteration mechanism implies that a single planar coordinate $(i, j)$ can theoretically generate up to $N$ distinct topological vertices, successfully capturing overlapping traversable spaces as demonstrated by the vertex generation on Plane 1 and Plane 2 in Fig.~\ref{fig6} (d) and (e).

Upon completing the comprehensive generation of vertices, the algorithm establishes edges to interconnect these topological vertices, incrementally updating $\mathcal{G}$ as shown in Fig.~\ref{fig6} (f). We construct a spatial partitioning tree $\mathcal{T}_{local}$ containing the comprehensive local vertex set $\mathcal{V}_{com} = \mathcal{V}_{new} \cup \mathcal{V}_{local}$. For a specific vertex $v \in \mathcal{V}_{com}$, a rapid neighborhood search is executed via $\mathcal{T}_{local}$ to identify all near vertices $\mathcal{N}_v$. To aggressively optimize subsequent graph search efficiency and prevent explosive computational overhead, we impose a strict mathematical upper bound $D_{max}$ on the allowable edge degree for every single vertex. Consequently, the algorithm strictly validates each candidate $n \in \mathcal{N}_v$ by evaluating two fundamental criteria. First, the number of edges already incident to both vertices must strictly satisfy $\text{degree}(v) < D_{max}$ and $\text{degree}(n) < D_{max}$. Second, the spatial connectivity must guarantee a completely collision-free linear trajectory, verified by Eq.~\ref{equation8} utilizing $\mathcal{M}^{L}_{grid}$. When both conditions are satisfied, a bidirectional topological edge $(v, n)$ is formally appended to the global edge set $\mathcal{E}$. The complete procedure for elevation-aware topological graph construction is summarized in \textbf{Algorithm 1}.

\subsection{Viewpoint Generation}
\label{Viewpoint Generation}

\begin{figure*}
	\begin{center}
		\includegraphics[width=\linewidth]{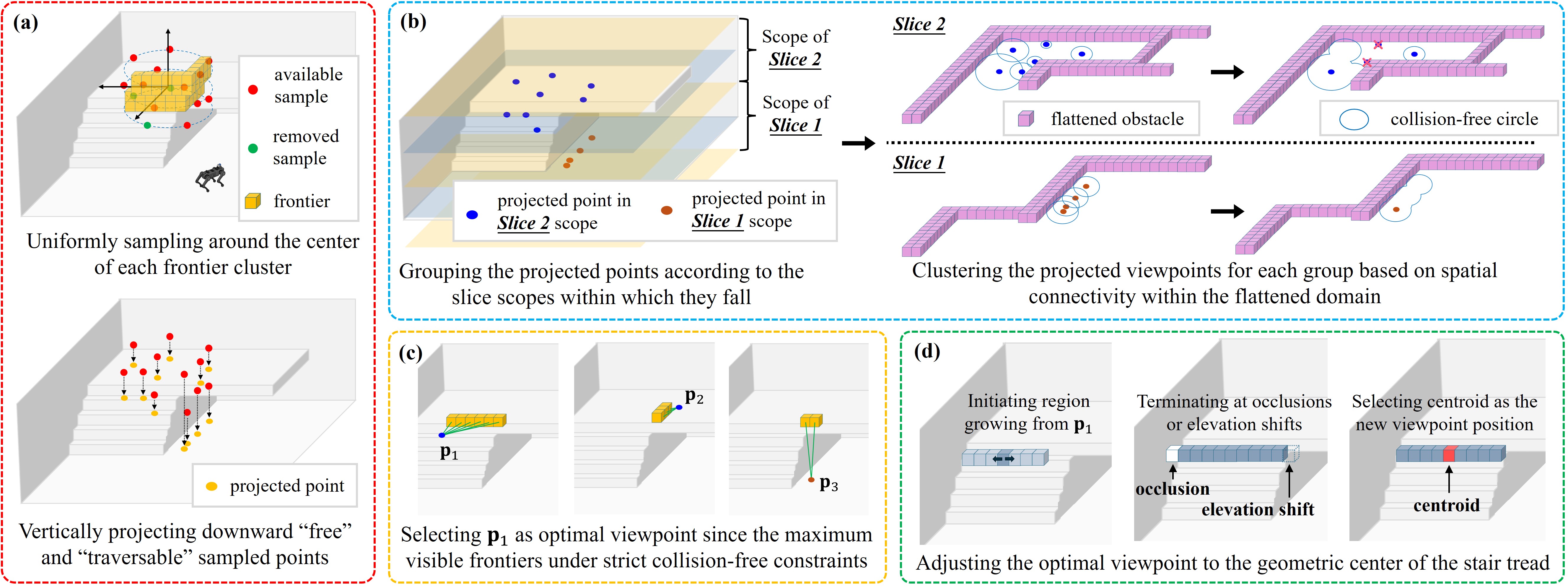}
		\caption{Illustration of viewpoint generation. (a) Raw samples are evaluated where invalid points (green) are discarded, and valid points (red) are vertically projected to surface-level positions (orange). (b) Projected points are categorized according to the specific slice scopes they fall into (e.g., blue and brown subsets). After eliminating points located too close to obstacles (red crosses), the remaining safe points are clustered using collision-free circles to extract candidate viewpoints. (c) The optimal viewpoint $\mathbf{p}_1$ is selected for maximizing frontier visibility compared to $\mathbf{p}_2$ and $\mathbf{p}_3$. (d) On staircases, an elevation-constrained region expansion from $\mathbf{p}_1$ extracts the tread centroid to serve as the new viewpoint position.}
		\label{fig7}
	\end{center}
\end{figure*}

Generating suitable viewpoints corresponding to newly identified frontiers is a critical step in autonomous exploration. A straightforward strategy is to uniformly sample candidate points around the robot and exhaustively execute collision-free ray casting against each frontier. This approach is computationally intensive, especially in sparsely occupied spaces where the large number of frontiers and candidate points leads to a considerable increase in computational cost. Inspired by EPIC \cite{geng2025epic}, we address this bottleneck by adopting a cluster-first strategy. We initially aggregate the raw frontiers into distinct spatial groups and subsequently generate and select an optimal viewpoint for each specific cluster, thereby restricting the computational complexity to a stable dimension. Furthermore, to prevent unsafe large yaw angle variations when the quadruped robot traverses staircases, we introduce a center alignment method that strictly adjusts any viewpoint located on a staircase to the geometric center of the stair tread.

\subsubsection{Frontier Clustering}
\label{Frontier Clustering}

Newly detected frontiers from $\mathcal{M}^{G}_{occ}$ are appended to queue $\mathcal{Q}_{f}$. Concurrently, any previously established frontier clusters that intersect with the map update region $\Omega_{upd}$ are removed, and their constituent frontiers are returned to $\mathcal{Q}_{f}$. All frontiers currently residing in $\mathcal{Q}_{f}$ are then reclustered based on their geometric proximity. Clusters containing an insufficient number of frontier cells are deemed insignificant and discarded, ultimately yielding the finalized set of newly aggregated frontier clusters $\mathcal{C}_{f}$.

\subsubsection{Candidate Viewpoint Generation}
\label{Candidate Viewpoint Generation}

As illustrated in Fig.~\ref{fig7} (a), the procedure initiates by uniformly sampling discrete spatial positions $\mathcal{P}_{raw}$ in a predefined cylindrical pattern around the centroid of each frontier cluster $h \in \mathcal{C}_{f}$. However, these initial coordinates might spatially float in the air or intersect with physical obstacles. To guarantee that these candidate locations are strictly situated on valid surfaces, we explicitly evaluate their structural status utilizing the local volumetric grid map $\mathcal{M}^{L}_{grid}$. Specifically, we eliminate any sampled point whose corresponding geometric state is evaluated as “occupied” or “unknown” (depicted as green points). For the surviving valid points (depicted as red points), we reassign their vertical coordinates as ground elevation $e^G_{i,j,k}$, where $(i, j)$ is calculated by Eq.~\ref{equation6} and $k$ is calculated by Eq.~\ref{equation10}. This vertical projection strategy yields a refined set of surface level projected points $\mathcal{P}_{proj}$ (depicted as orange points), where each point $\mathbf{p} \in \mathcal{P}_{proj}$ is formally instantiated as $[[\mathbf{p}]_x, [\mathbf{p}]_y, e^G_{i,j,k}]^T$.

Following this projection, as shown in Fig.~\ref{fig7} (b), the algorithm systematically categorizes these projected points into distinct subsets $\mathcal{P}_{proj}^k$ depending on the specific tomogram slice scope $k$ within which they vertically reside (e.g., the blue and brown groups). Subsequently, within the flattened spatial domain of each distinct group, we integrate the slice-wise obstacles extracted from Sec.~\ref{Slice-Based Obstacle Extraction}. We first rigorously filter out any projected point that fails to maintain a safe clearance $d_{safe}$ from nearby obstacles utilizing Eq.~\ref{equation11} (depicted as red crosses). For the remaining structurally safe projected points, we execute a connectivity based clustering procedure to form robust candidate viewpoints. Specifically, for each point $\mathbf{p} \in \mathcal{P}_{proj}^k$, we utilize the result from Eq.~\ref{equation11} to construct a collision-free circle whose radius equates to its shortest spatial distance to the nearest obstacle. If the collision-free circles of two projected points geometrically intersect, they are merged into the same spatial cluster. Ultimately, within each formed cluster, the specific projected point possessing the maximum obstacle distance is selected as the representative candidate viewpoint. All successfully extracted representatives are collected into the comprehensive candidate viewpoint set $\mathcal{P}_{cand}$. The complete procedure for candidate viewpoint generation is summarized in \textbf{Algorithm 2}.

\begin{algorithm}[t]
	\caption{Candidate Viewpoint Generation}
	\label{alg:candidate_viewpoint}
	\small
	\begin{algorithmic}[1]
		\Input Frontier cluster centroid $\mathbf{p}_{c}$, local tomography map $\mathcal{M}^{L}_{tomo}$, local 3D grid map $\mathcal{M}^{L}_{grid}$, vertical local map boundary minimum $z_{min}$, slice-wise obstacle trees $\mathcal{T}_{obs} = \{\mathcal{T}_{obs}^k\}$
		\Output Set of candidate viewpoints $\mathcal{P}_{cand}$
		\Hyperparameters Vertical interval of adjacent slices $d_s$, safety margin to obstacles $d_{safe}$
		\State $\mathcal{P}_{cand} \leftarrow \emptyset$
		\State $\mathcal{P}_{raw} \leftarrow$ uniformly sampling around $\mathbf{p}_{c}$
		\State $\mathcal{P}_{proj} \leftarrow \emptyset$
		\State 
		\State \textbf{Phase 1: Valid Surface Projection}
		\For{each raw point $\mathbf{p} \in \mathcal{P}_{raw}$}
		\State $k = \lfloor ([\mathbf{p}]_z - z_{min}) / d_s \rfloor$
		\State $\sigma_{\mathbf{p}} = \text{getGridState}(\mathbf{p})$
		\If{$\sigma_{\mathbf{p}} \neq \sigma_{occ}$ \textbf{and} $\sigma_{\mathbf{p}} \neq \sigma_{unk}$}
		\If{ground elevation $e^G_{i,j,k}$ is valid}
		\State $\mathbf{p}_{proj} \leftarrow [[\mathbf{p}]_x, [\mathbf{p}]_y, e^G_{i,j,k}]^T$
		\State $\mathcal{P}_{proj} \leftarrow \mathcal{P}_{proj} \cup \{\mathbf{p}_{proj}\}$
		\EndIf
		\EndIf
		\EndFor
		\State 
		\State \textbf{Phase 2: Slice Grouping and Safe Clustering}
		\State Group $\mathcal{P}_{proj}$ into slice subsets $\{\mathcal{P}_{proj}^k\}$ based on elevation scopes
		\For{each slice index $k$}
		\State $\mathcal{P}_{safe}^k \leftarrow \emptyset$
		\For{each projected point $\mathbf{p} \in \mathcal{P}_{proj}^k$}
		\State $d_{obs} \leftarrow \text{getObstacleDistance}(\mathcal{T}_{obs}^k, [\mathbf{p}]_{x,y})$
		\If{$d_{obs} > d_{safe}$}
		\State $rad_{\mathbf{p}} \leftarrow d_{obs}$
		\State $\mathcal{P}_{safe}^k \leftarrow \mathcal{P}_{safe}^k \cup \{\mathbf{p}\}$
		\EndIf
		\EndFor
		\State 
		\State Initialize spatial cluster set $\mathcal{C}_{cand} \leftarrow \emptyset$
		\State Cluster $\mathcal{P}_{safe}^k$ into $\mathcal{C}_{cand}$ merging points if $\Vert{}\mathbf{p}_a - \mathbf{p}_b\Vert{} \le rad_{\mathbf{p}_a} + rad_{\mathbf{p}_b}$
		\For{each cluster $\beta \in \mathcal{C}_{cand}$}
		\State Extract representative $\mathbf{p}_{cand} \leftarrow \underset{\mathbf{p} \in \beta}{\arg\max} \, (rad_{\mathbf{p}})$
		\State $\mathcal{P}_{cand} \leftarrow \mathcal{P}_{cand} \cup \{\mathbf{p}_{cand}\}$
		\EndFor
		\EndFor
		\State 
		\Return $\mathcal{P}_{cand}$
	\end{algorithmic}
\end{algorithm}

\subsubsection{Optimal Viewpoint Selection}
\label{Optimal Viewpoint Selection}

We evaluate the exploration efficiency of each candidate viewpoint in $\mathcal{P}_{cand}$ by quantifying the number of visible frontiers within the corresponding frontier cluster $h$. To precisely determine this visibility count, we execute collision-free line of sight checks utilizing Eq. 8 between every candidate viewpoint and all frontiers within $h$. As exemplified in Fig.~\ref{fig7} (c), the candidate $\mathbf{p}_1$ successfully observes a maximum number of frontiers compared to the alternative candidates $\mathbf{p}_2$ and $\mathbf{p}_3$. Consequently, $\mathbf{p}_1$ is selected as the optimal viewpoint for the cluster $h$.

\subsubsection{Center Alignment for Staircases}
\label{Center Alignment for Staircases}

Although the selected optimal viewpoint $\mathbf{p}_1$ guarantees visibility, its exact placement on narrow structures such as stairs might locate perilously close to the step edges, posing a severe locomotion risk for the quadruped platform. We mathematically identify the presence of a staircase terrain by evaluating the structural elevation deviation. Specifically, if the absolute height difference between the current robot position $\mathbf{p}_{curr}$ and the optimal viewpoint $\mathbf{p}_1$ strictly exceeds the map grid resolution $r_{grid}$, the local terrain is classified as a staircase. Under this specific condition, we initiate a center alignment mechanism to adjust the optimal viewpoint towards the geometric center of the stair tread.

As explicitly illustrated in Fig.~\ref{fig7} (d), we formulate the center alignment as an iterative spatial aggregation process originating from the seed grid $w_{ini}$ where $\mathbf{p}_1$ locates. We define $\mathcal{C}_{tread}$ as the equipotential spatial cluster for representing the contiguous stair tread, initially defined as $\mathcal{C}_{tread} = \{w_{ini}\}$. During the region growing procedure, we systematically explore the local neighborhood $\mathcal{N}_w$ for every grid $w \in \mathcal{C}_{tread}$. An adjacent candidate grid $w_{adj} \in \mathcal{N}_w$ is absorbed into $\mathcal{C}_{tread}$ if and strictly if it satisfies two geometric conditions: it must be structurally unoccupied, and its ground elevation $e^G_{w_{adj}}$ must adhere to the coplanar constraint $\vert{}e^G_{w_{adj}} - e^G_{w_{ini}}\vert{} < r_{grid}$, where $r_{grid}$ is the resolution of $\mathcal{M}^{L}_{grid}$, serving here as the maximum allowable elevation tolerance. This outward propagation proceeds continuously, provided that the above two geometric constraints are satisfied, until the maximum cluster cardinality $\gamma$ is reached. Ultimately, to derive the newly refined and optimally safe viewpoint position $\mathbf{p}_{opt}$, we calculate the geometric centroid of the converged bounded cluster as follows:
\begin{equation}
	\label{equation13}
	\small
	\mathbf{p}_{opt} = \frac{1}{\vert{}\mathcal{C}_{tread}\vert{}} \sum_{\mathbf{p} \in \mathcal{C}_{tread}} \mathbf{p}
\end{equation}

\subsection{Dual Path Searching}
\label{Dual Path Searching}

Following the generation of the new viewpoint set $\mathcal{P}_{vp}$, the algorithm must resolve the optimal visiting sequence for viewpoints. We temporarily connect every viewpoint within $\mathcal{P}_{vp}$ to its spatially adjacent topological vertices in $\mathcal{G}$, provided that the direct spatial linkage between them strictly satisfies the collision-free verification formulated in Eq.~\ref{equation8}. We then systematically search the path from the current robot position to each connected viewpoint, alongside the pairwise paths between all viewpoints, ultimately formulating a comprehensive cost matrix.

\begin{figure}
	\begin{center}
		\includegraphics[width=\linewidth]{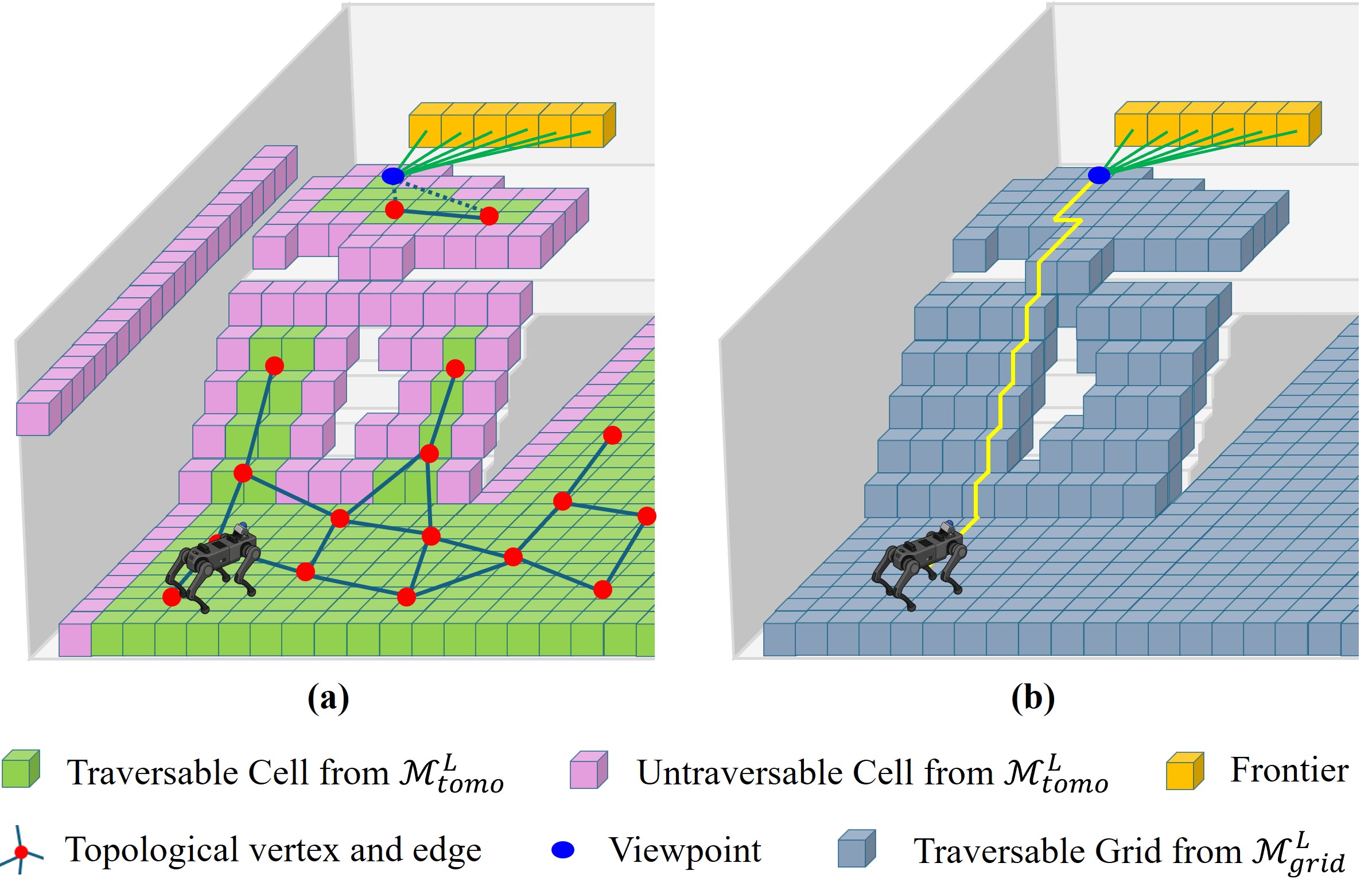}
		\caption{Illustration of the dual path searching approach for staircases. (a) Graph-based searching fails due to topological discontinuity. (b) The fallback traversable-grid based searching successfully resolves a valid trajectory (yellow path) to the upper-floor viewpoint.}
		\label{fig8}
	\end{center}
\end{figure}

As illustrated in Fig.~\ref{fig8} (a), the sensory point clouds captured in staircase regions are frequently sparse or fragmented, inherently yielding degraded local tomogram representations. Consequently, the spatial distance between the topological vertices distributed along the staircase and those situated on the upper-floor landing frequently exceeds the maximum allowable connection threshold. This geometric isolation prevents direct edge formulation, leading to topological discontinuity that inevitably causes graph-based A-star search to fail during cross-floor operations.

To mitigate this critical vulnerability, we propose a robust dual path searching approach in this section. In this dual strategy, we first attempt to search a path relying on the established $\mathcal{G}$. If this primary search fails due to a topological disconnection, the algorithm executes a rigorous grid-based A-star search. This dense path searching is exclusively constrained to spatial grids explicitly labeled as “traversable” within the discrete state space of $\mathcal{M}^{L}_{grid}$. Recognizing that repetitive volumetric path planning exponentially increases the computational burden during the cost matrix formulation, we engineer a bidirectional path caching mechanism. Whenever a successful path is searched, both the forward and reversed geometric paths are symmetrically preserved in a persistent memory matrix. Consequently, subsequent identical pairwise queries instantaneously retrieve the cached spatial trajectory. As illustrated in Fig.~\ref{fig8} (b), the yellow path represents the valid path searched by the traversable grid-based A-star search, successfully evaluating the motion cost from the current robot position to the upper-floor viewpoint. This success is primarily attributed to the more lenient classification criteria for traversable grids within $\mathcal{M}^{L}_{grid}$ compared to the strict traversable cells within $\mathcal{M}^{L}_{tomo}$. This dual path searching approach significantly amplifies the reliability of motion cost evaluation in staircases. By effectively preventing path searching failures, it guarantees the uninterrupted generation of upward guidance trajectories, thereby ensuring the quadruped robot can successfully and continuously climb multi-floor structures. Finally, for any viewpoint where both stages of the dual path search fail (as illustrated in Fig.~\ref{fig1} (d)) it is classified as strictly unreachable. Such isolated viewpoints are explicitly excluded from the cost matrix formulation and are consequently ignored during the subsequent computation of optimal visitation sequences.

\begin{table}[]
	\caption{Information Of Simulation Scenarios}
	\label{table2}
	\centering
	\begin{tabular}{c|ccc}
		\toprule
		& \textbf{Size (m)}      & \textbf{Floor Count} & \textbf{Inter-Floor Height (m)} \\ \hline
		$scene\_1$ & 18.4$\times$18.4$\times$8.2 & 3           & 2.1                \\
		$scene\_2$ & 13.0$\times$22.4$\times$8.2 & 3           & 2.5                \\
		$scene\_3$ & 24.0$\times$34.0$\times$7.7 & 2           & 3.4                \\
		$scene\_4$ & 17.6$\times$16.8$\times$6.2 & 4           & 1.5                \\ \bottomrule
	\end{tabular}
\end{table}

\begin{table}[]
	\caption{Parameter Configurations}
	\label{table3}
	\centering
	\begin{tabular}{p{0.6cm}<{\centering}|p{0.9cm}<{\centering}p{0.9cm}<{\centering}p{0.9cm}<{\centering}p{0.9cm}<{\centering}c}
		\toprule
		& \multicolumn{1}{c|}{\textbf{scene\_1}} & \multicolumn{1}{c|}{\textbf{scene\_2}} & \multicolumn{1}{c|}{\textbf{scene\_3}} & \multicolumn{1}{c|}{\textbf{scene\_4}} & \textbf{Real-World} \\ \hline
		$R_{occ}$	 & \multicolumn{4}{c|}{4.0}                                                                                 & 4.0        \\
		$\mathcal{B}_{local}$	& \multicolumn{1}{c|}{12$\times$12$\times$4}       & \multicolumn{1}{c|}{12$\times$12$\times$6}       & \multicolumn{2}{c|}{12$\times$12$\times$4}                  & 12$\times$12$\times$6         \\
		$N$  & \multicolumn{4}{c|}{5}                                                                                 & 5        \\
		$r_{tomo}$  & \multicolumn{4}{c|}{0.2}                                                                                 & 0.2        \\
		$d_{min}$   & \multicolumn{4}{c|}{0.5}                                                                                 & 0.5        \\
		$\theta_s$  & \multicolumn{4}{c|}{0.6}                                                                                 & 0.6        \\
		$\theta_b$  & \multicolumn{4}{c|}{0.3}                                                                                 & 0.3        \\
		$\theta_p$  & \multicolumn{1}{c|}{0.5}       & \multicolumn{1}{c|}{0.1}       & \multicolumn{2}{c|}{0.5}                  & 0.1         \\
		$c^B$       & \multicolumn{4}{c|}{50}                                                                                  & 50         \\
		$\alpha_d$  & \multicolumn{4}{c|}{20}                                                                                  & 20         \\
		$\alpha_s$  & \multicolumn{4}{c|}{15}                                                                                  & 15         \\
		$\alpha_b$  & \multicolumn{4}{c|}{20}                                                                                  & 20         \\
		$r_{grid}$  & \multicolumn{4}{c|}{0.2}                                                                                 & 0.2        \\
		$R_{topo}$  & \multicolumn{4}{c|}{6}                                                                                   & 6          \\
		$d_{safe}$  & \multicolumn{4}{c|}{0.4}                                                                                 & 0.4        \\
		$r_{min}$   & \multicolumn{4}{c|}{0.5}                                                                                 & 0.5        \\
		$D_{max}$   & \multicolumn{4}{c|}{6}                                                                                   & 6          \\
		$\gamma$    & \multicolumn{1}{c|}{50}       & \multicolumn{1}{c|}{16}       & \multicolumn{2}{c|}{50}                  & 16         \\
		$\lambda_z$ & \multicolumn{4}{c|}{100}                                                                                 & 100        \\
		$\mu_z$     & \multicolumn{4}{c|}{25}                                                                                  & 25        \\ \bottomrule
	\end{tabular}
\end{table}

\subsection{Visitation Sequence Solving}
\label{Visitation Sequence Solving}

Once the motion cost is evaluated, the exploration sequence is formulated as an Asymmetric Traveling Salesman Problem (ATSP) \cite{applegate2011traveling}. We define index $0$ as the current robot position and $\{1, \dots, E\}$ as the candidate viewpoints. The basic transition weight $\omega(a, b)$ ($a, b \in \{0, 1, \dots, E\}$) relies on the trajectory length $L(a, b)$ from path searching. To ensure the exhaustive exploration of the current floor before moving to the upper layer, we introduce an elevation penalty factor $\rho(a, b)$:
\begin{equation}
	\label{equation14}
	\small
	\omega(a, b) = L(a, b) \cdot \rho(a, b)
\end{equation}
This factor is activated strictly when the absolute height difference $\Delta z_{a,b}$ exceeds the local 3D grid map resolution $r_{grid}$:
\begin{equation}
	\label{equation15}
	\small
	\rho(a, b) = \begin{cases} 1 + \lambda_z (1 - \exp(-\mu_z \Delta z_{a,b})) & \text{if } \Delta z_{a,b} \ge r_{grid} \\ 1 & \text{otherwise} \end{cases}
\end{equation}
where $\lambda_z$ represents the maximum allowable penalty and $\mu_z$ regulates the exponential growth rate. Additionally, the initial path segment $\omega(0, b)$ incorporates a kinematic yaw penalty to suppress drastic in place rotations. To prevent circular closed loop trajectories, the return weight is zeroed out as $\omega(a, 0) = 0$. Ultimately, a two opt heuristic algorithm \cite{croes1958method} balances these penalized weights against information gains to extract an efficient and kinematically stable visitation sequence. The evaluation criterion for the information gain of each viewpoint is strictly identical to FAEL \cite{huang2023fael}, to which readers are referred for detailed formulations.

\subsection{Local Planner}
\label{Local Planner}

The local planner is a targeted modification of Ego-Planner \cite{zhou2020ego}, which was originally designed for UAVs. To adapt this aerial planner for a ground based quadruped robot, we introduce two primary improvements. First, an explicit penalty factor along the vertical $z$ axis is added during trajectory optimization. This constraint forces the generated paths to closely adhere to the physical ground surface, preventing physically infeasible floating trajectories. Second, the collision avoidance representation is upgraded from a global terrain map to a rolling window-based local inflated 3D grid map $\mathcal{M}^{L}_{ter}$. This localized structure maintains a constant memory overhead while maintaining high fidelity obstacle awareness in the immediate vicinity of the robot.

\section{Experiments}
\label{Experiments}

In this section, we conduct extensive experiments in both simulation and real-world environments to evaluate the proposed exploration framework RAEM. The quantitative comparisons with other ground robot exploration frameworks, execution time evaluations, and ablation studies are exclusively conducted in the simulation environment. Conversely, the real-world experiments are primarily utilized to verify that the proposed framework operates robustly in actual scenarios characterized by map noise and imperfect localization accuracy.

All simulation experiments are evaluated within the Gazebo simulator, running on a laptop equipped with an Intel Core i7 12700H CPU, a GeForce RTX 3050Ti GPU, and 32\,GB of RAM. We employ a Unitree A1 quadruped as the robotic platform, equipped with a Mid-360 LiDAR ($360^\circ \times 59^\circ$ FOV). This LiDAR sensor is installed on the head of the robot at a pitch angle of $45^\circ$, and its effective perception range is configured to 4.0\,m. We have customized four distinct simulation scenarios. $scene\_1$ is a three-floor environment featuring a hollow square staircase. $scene\_2$ is a three-floor building where adjacent floors are connected by corner staircases. $scene\_3$ represents a two-floor building containing multiple individual rooms on each layer. Finally, $scene\_4$ is a highly complex four-floor scenario incorporating narrow corridors, asymmetrically positioned staircases, blind corners situated behind stairs, and regions that are physically traversable but spatially unreachable due to structural obstructions (as shown in Fig.~\ref{fig1}). Detailed information regarding each scenario, including geometric sizes, number of floors, and inter-floor spacing, is summarized in Table~\ref{table2}.

\begin{figure}[t!]
	\begin{center}
		\includegraphics[width=\linewidth]{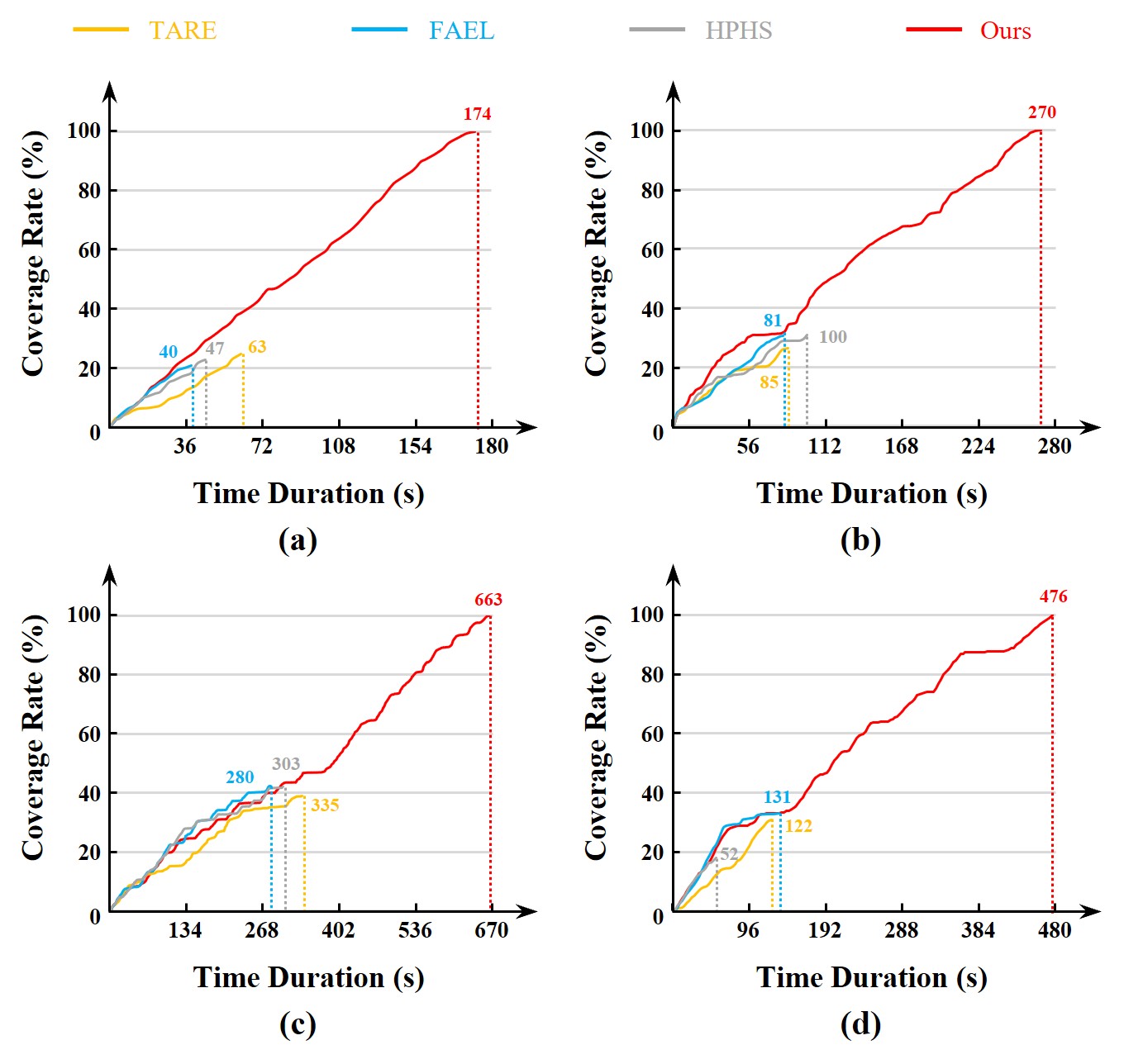}
		\caption{The variation of coverage rate with time duration in exploration progress for (a) $scene\_1$, (b) $scene\_2$, (c) $scene\_3$, and (d) $scene\_4$.}
		\label{fig9}
	\end{center}
\end{figure}

\begin{figure}[t!]
	\begin{center}
		\includegraphics[width=\linewidth]{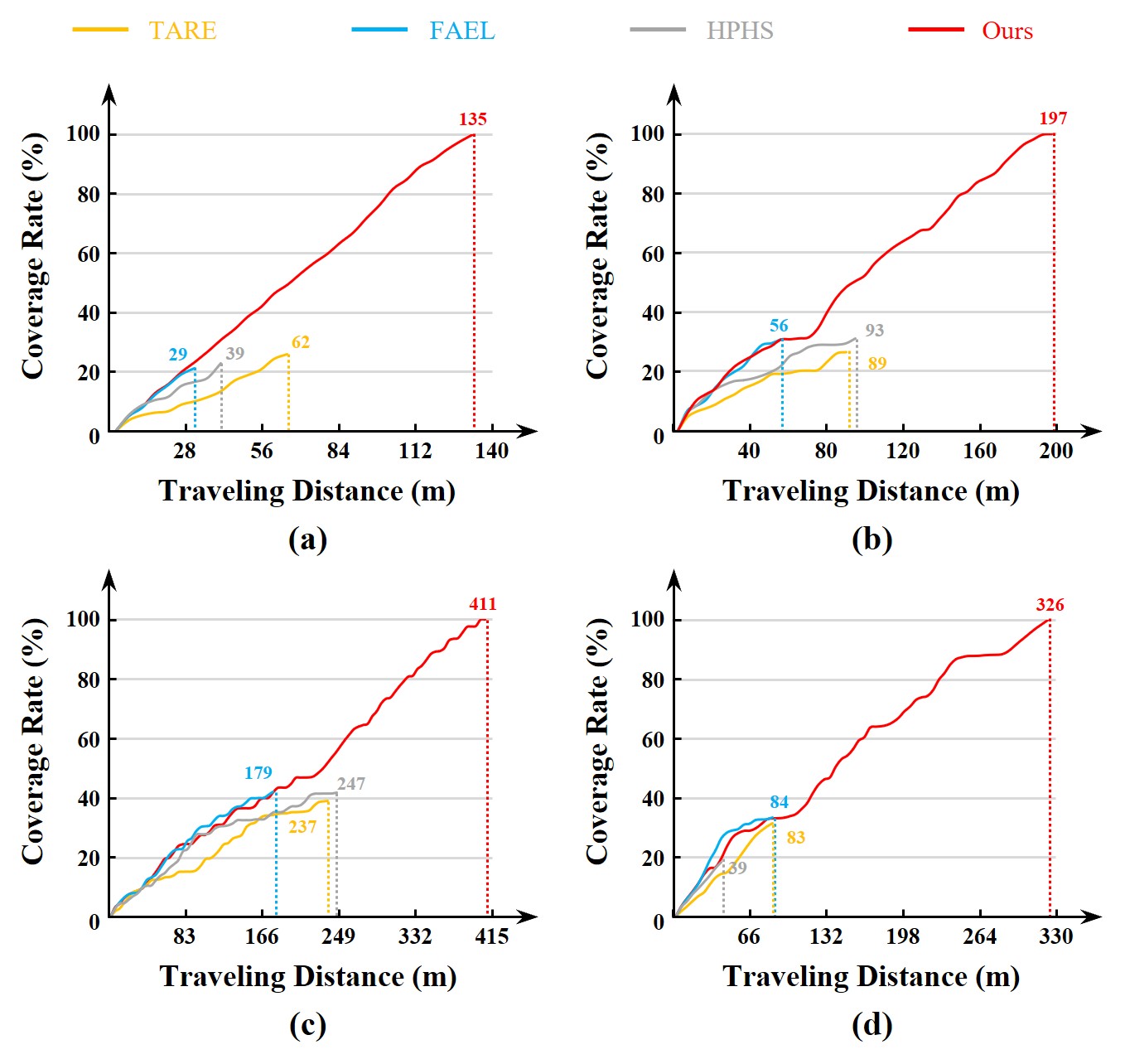}
		\caption{The variation of coverage rate with traveling distance in exploration progress for (a) $scene\_1$, (b) $scene\_2$, (c) $scene\_3$, and (d) $scene\_4$.}
		\label{fig10}
	\end{center}
\end{figure}

The specific hardware configurations for the real-world experiments are introduced in Sec.~\ref{Real-World Experiment}. Table~\ref{table3} presents the specific assigned values for all hyperparameters across both the simulation and real-world scenarios. Except for a few specific variables, the vast majority of these parameter values remain consistent across all experimental setups.

\subsection{Comparison with State-of-the-Arts}
\label{Comparison with State-of-the-Arts}

\begin{figure*}
	\begin{center}
		\includegraphics[width=\textwidth]{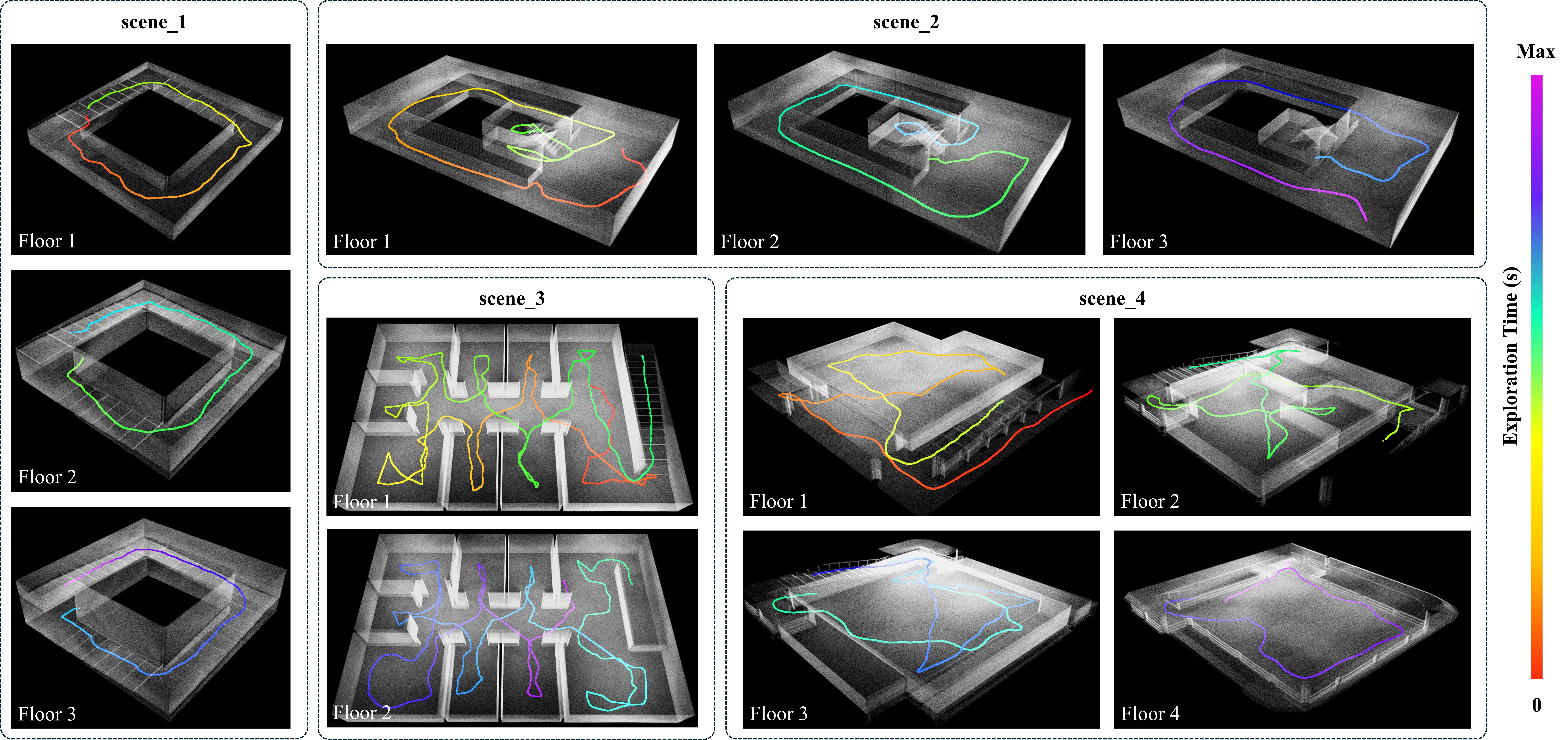}
		\caption{Visualization of exploration paths generated by RAEM across $scene\_1$, $scene\_2$, $scene\_3$, and $scene\_4$.}
		\label{fig11}
	\end{center}
\end{figure*}

\begin{table*}[]
	\caption{Results of Simulations in four Environments}
	\label{table4}
	\centering
	\begin{threeparttable}
	\begin{tabular}{c|c|cccc|cccc|cccc|cccc}
		\toprule
		&      & \multicolumn{4}{c|}{\textbf{Exploration Time (s)}}          & \multicolumn{4}{c|}{\textbf{Trajectory Distance (m)}}         & \multicolumn{4}{c|}{\textbf{Occupied Volume (m$^3$)}} & \multicolumn{4}{c}{\textbf{Success Number}}                                                                                                                                                                                                               \\ \hline
		&      & \textbf{Avg}   & \textbf{Std} & \textbf{Max} & \textbf{Min} & \textbf{Avg}   & \textbf{Std} & \textbf{Max} & \textbf{Min} & \textbf{Avg}      & \textbf{Std}   & \textbf{Max}   & \textbf{Min}   & \textbf{\begin{tabular}[c]{@{}c@{}}1st\\ Floor\end{tabular}} & \textbf{\begin{tabular}[c]{@{}c@{}}2nd\\ Floor\end{tabular}} & \textbf{\begin{tabular}[c]{@{}c@{}}3rd\\ Floor\end{tabular}} & \textbf{\begin{tabular}[c]{@{}c@{}}4th\\ Floor\end{tabular}} \\ \hline
		\multirow{4}{*}{$scene\_1$} & TARE & 63           & 2          & 67         & 59         & 62           & 1          & 65         & 60         & 102             & 3            & 107          & 95           & 20                                                           & 0                                                            & 0                                                            &                                                             \\
		& FAEL & 41           & 5          & 48         & 32         & 32           & 4          & 37         & 26         & 90              & 11           & 109          & 74           & 20                                                           & 0                                                            & 0                                                            &                                                             \\
		& HPHS & 41           & 2          & 47         & 37         & 39           & 2          & 46         & 35         & 89              & 1            & 90           & 87           & 20                                                           & 0                                                            & 0                                                            &                                                             \\
		& Ours & \textbf{174} & 3          & 184        & 166        & \textbf{137} & 2          & 141        & 134        & \textbf{389}    & 4            & 397          & 379          & 20                                                           & \textbf{20}                                                  & \textbf{20}                                                  &                                                   \\ \hline
		\multirow{4}{*}{$scene\_2$} & TARE & 86           & 9          & 108        & 74         & 81           & 8          & 104        & 72         & 109             & 6            & 122          & 100          & 20                                                           & 0                                                            & 0                                                            &                                                             \\
		& FAEL & 82           & 6          & 96         & 72         & 59           & 3          & 67         & 54         & 121             & 1            & 122          & 118          & 20                                                           & 0                                                            & 0                                                            &                                                             \\
		& HPHS & 110          & 26         & 141        & 81         & 101          & 23         & 129        & 76         & 120             & 1            & 122          & 119          & 5                                                            & 0                                                            & 0                                                            &                                                             \\
		& Ours & \textbf{270} & 11         & 290        & 246        & \textbf{195} & 4          & 205        & 188        & \textbf{393}    & 2            & 396          & 390            & 20                                                           & \textbf{20}                                                  & \textbf{20}                                                  &                                                   \\ \hline
		\multirow{4}{*}{$scene\_3$} & TARE & 333          & 20         & 358        & 304        & 235          & 17         & 258        & 214          & 421             & 13           & 442          & 408          & 20                                                           & 0                                                            &                                                             &                                                             \\
		& FAEL & 281          & 16         & 305        & 259        & 171          & 7          & 185        & 160        & 441             & 8            & 455          & 428          & 20                                                           & 0                                                            &                                                             &                                                             \\
		& HPHS & 305          & 23         & 341        & 269        & 255          & 17         & 286        & 228        & 442             & 3            & 448          & 437          & 7                                                            & 0                                                            &                                                             &                                                             \\
		& Ours & \textbf{670} & 25         & 724        & 628        & \textbf{416} & 11         & 442        & 398        & \textbf{1065}   & 10            & 1088         & 1044         & 20                                                           & \textbf{20}                                                  &                                                   &                                                   \\ \hline
		\multirow{4}{*}{$scene\_4$} & TARE & 122          & 12         & 158        & 96         & 84           & 6          & 94         & 68         & 155             & 7            & 165          & 141          & 20                                                           & 0                                                            & 0                                                            & 0                                                            \\
		& FAEL & 130          & 10          & 147        & 111        & 87           & 6          & 97         & 74         & 164             & 3            & 170          & 160          & 20                                                           & 0                                                            & 0                                                            & 0                                                            \\
		& HPHS & -              & -            & -            & -            & -              & -            & -            & -            & -                 & -              & -              & -              & 0                                                            & 0                                                            & 0                                                            & 0                                                            \\
		& Ours & \textbf{477} & 30         & 550        & 404        & \textbf{322} & 15         & 352        & 285        & \textbf{501}    & 8            & 519          & 484          & 20                                                           & \textbf{20}                                                  & \textbf{20}                                                  & \textbf{20}                                                  \\ \bottomrule
	\end{tabular}
\end{threeparttable}
\begin{tablenotes}
	\footnotesize
	\item[] \textbf{Denotations}: “-” indicates that the method failed to successfully explore at least one complete floor in any trial, making this value unrecordable.
	\end{tablenotes}
\end{table*}

\begin{figure*}
	\begin{center}
		\includegraphics[width=\linewidth]{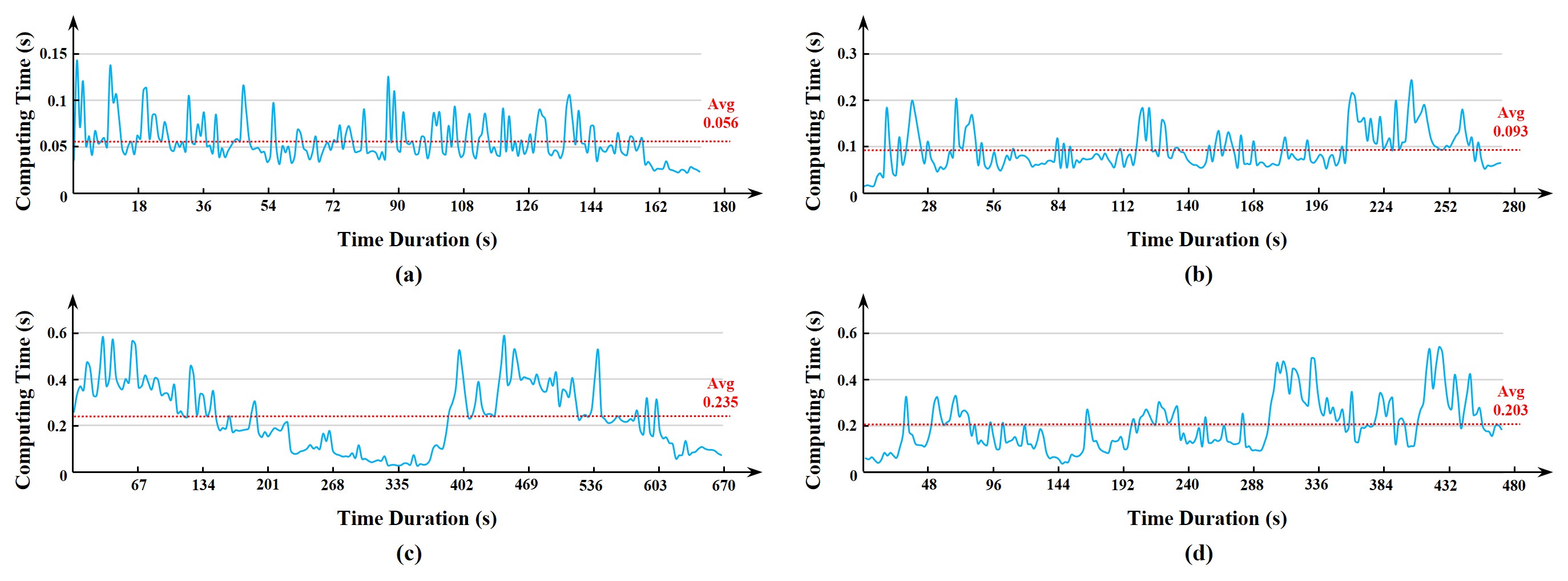}
		\caption{Evolution of the computation time for a single planning iteration as time progresses in (a) $scene\_1$, (b) $scene\_2$, (c) $scene\_3$, and (d) $scene\_4$.}
		\label{fig12}
	\end{center}
\end{figure*}

\begin{figure*}
	\begin{center}
		\includegraphics[width=\linewidth]{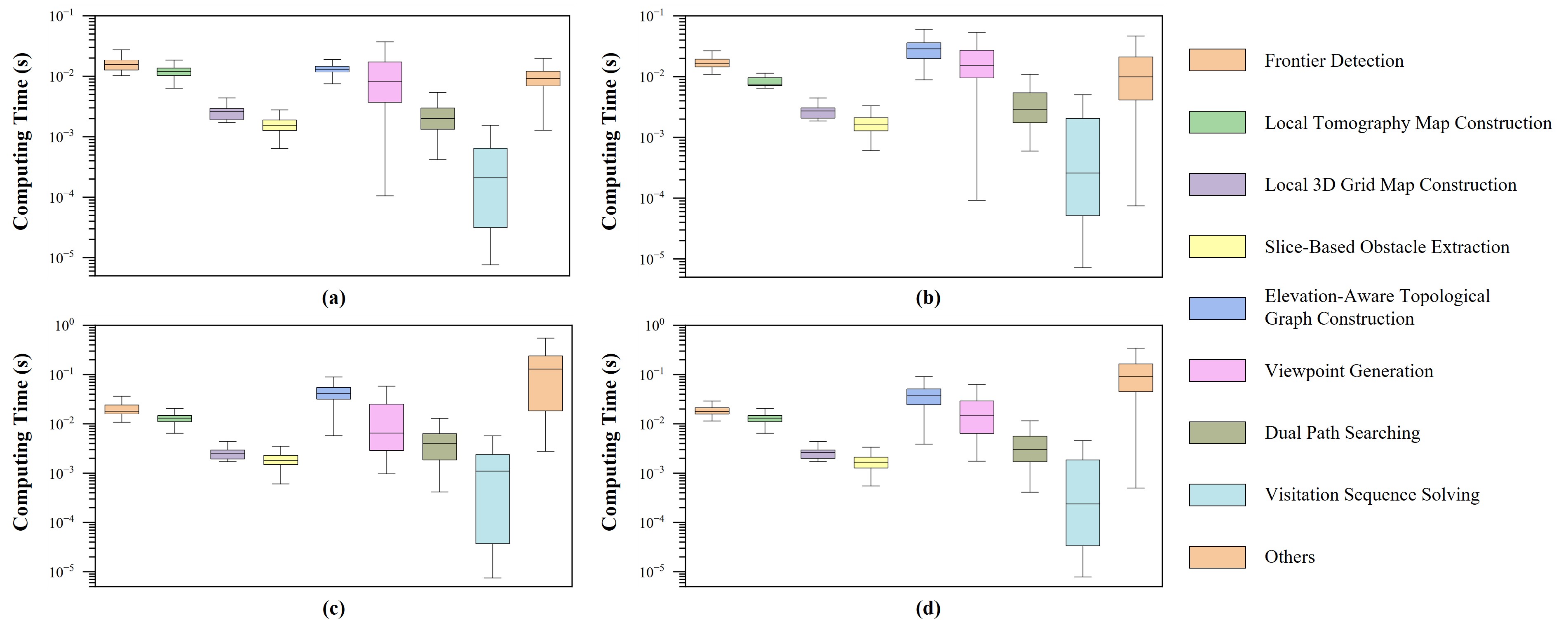}
		\caption{Box plot with logarithmic scale of single computation time of each module for (a) $scene\_1$, (b) $scene\_2$, (c) $scene\_3$, and (d) $scene\_4$.}
		\label{fig13}
	\end{center}
\end{figure*}

We compare our method with three representative LiDAR-based ground exploration frameworks, i.e., TARE \cite{cao2021tare}, FAEL \cite{huang2023fael} and HPHS \cite{long2024hphs}. To ensure a fair evaluation, we uniformly set the grid resolution for all approaches to 0.2\,m and configure the maximum velocity of quadruped robot to 1.0\,m/s, which is strictly consistent with our framework. Furthermore, since HPHS operates fundamentally as a 2D grid-based method and inherently lacks 3D spatial elements, we artificially construct a supplementary 3D grid data structure specifically to evaluate the explored occupied volume. Each method was performed 20 times in each environment at the same start position, and the quantitative evaluation results are provided in Table~\ref{table4}. Fig.~\ref{fig9} and Fig.~\ref{fig10} illustrate the exploration progress, represented by the single trial whose exploration time result was closest to the average, for all four approaches in the four simulation environments. The metric termed "coverage rate" here refers to the percentage of the occupied volume relative to the final total occupied volume successfully explored by RAEM. Furthermore, we present the visualization of the exploration paths generated by RAEM across the four simulation scenarios, as illustrated in Fig.~\ref{fig11}. Since statically visualizing the continuous trajectory across an entire multi-floor structure within a single image inherently causes severe visual clutter, we adopt a layer-by-layer visualization strategy to independently display the exploration paths for each distinct floor. Comprehensive demonstration of the dynamic exploration process can be found in supplementary video.

From $scene\_1$ to $scene\_4$, both TARE and FAEL successfully complete the exploration of the first floor. However, due to the lack of traversability representation capabilities designed for multi-floor structures, the quadruped robot is unable to navigate toward higher layers using these frameworks. Furthermore, HPHS fails to stably achieve a complete exploration of the first floor in $scene\_2$ and $scene\_3$, and it does not record a single successful attempt in $scene\_4$. This systematic failure arises because HPHS does not provide a guidance path to the local planner module; rather, it solely transmits the isolated target waypoint, completely offloading the entire trajectory generation burden onto the local planner. When navigating structurally complex or severely narrow environments, relying exclusively on a local planner for spatial pathfinding is highly susceptible to failure. In contrast to the three aforementioned approaches, the proposed framework accomplishes stable and exhaustive exploration across 20 independent test trials.

\subsection{Computational Cost}
\label{Computational Cost}

To comprehensively evaluate the computational overhead of RAEM, we plot curves illustrating the computation time for a single planning iteration as time progresses across the four simulation scenarios in Fig.~\ref{fig12}. In $scene\_1$ and $scene\_2$, since the internal spatial layouts predominantly consist of corridors and staircases, both the number of detected frontiers and generated viewpoints remain distinctly limited. Consequently, the average computational cost of a single planning iteration falls below 0.1\,s. Conversely, $scene\_3$ and $scene\_4$ contain extensive sparsely occupied spaces. These regions yield a large number of frontiers and viewpoints, thereby elevating the average computational expenditure of a single planning iteration to more than 0.2\,s. Furthermore, we evaluate the single execution time for all constituent modules across $scene\_1$ through $scene\_4$ utilizing the box plot format, as illustrated in Fig.~\ref{fig13}. It is readily apparent that the primary computational burdens are concentrated within the elevation-aware topological construction, the viewpoint generation, and others (which includes registering point clouds into the UFOMap \cite{duberg2020ufomap} and executing a coarse preliminary sorting of all candidate viewpoints prior to path planning). Notably, the computing time for “others” in $scene\_3$ and $scene\_4$ is significantly higher than that in $scene\_1$ and $scene\_2$. This increase occurs as sparsely occupied spaces inevitably generate a larger volume of viewpoints, which consequently expands the operational scale of the sorting algorithm.

\subsection{Ablation Study of GPU Acceleration}
\label{Ablation Study of GPU Acceleration}

\begin{figure}[t]
	\begin{center}
		\includegraphics[width=\linewidth]{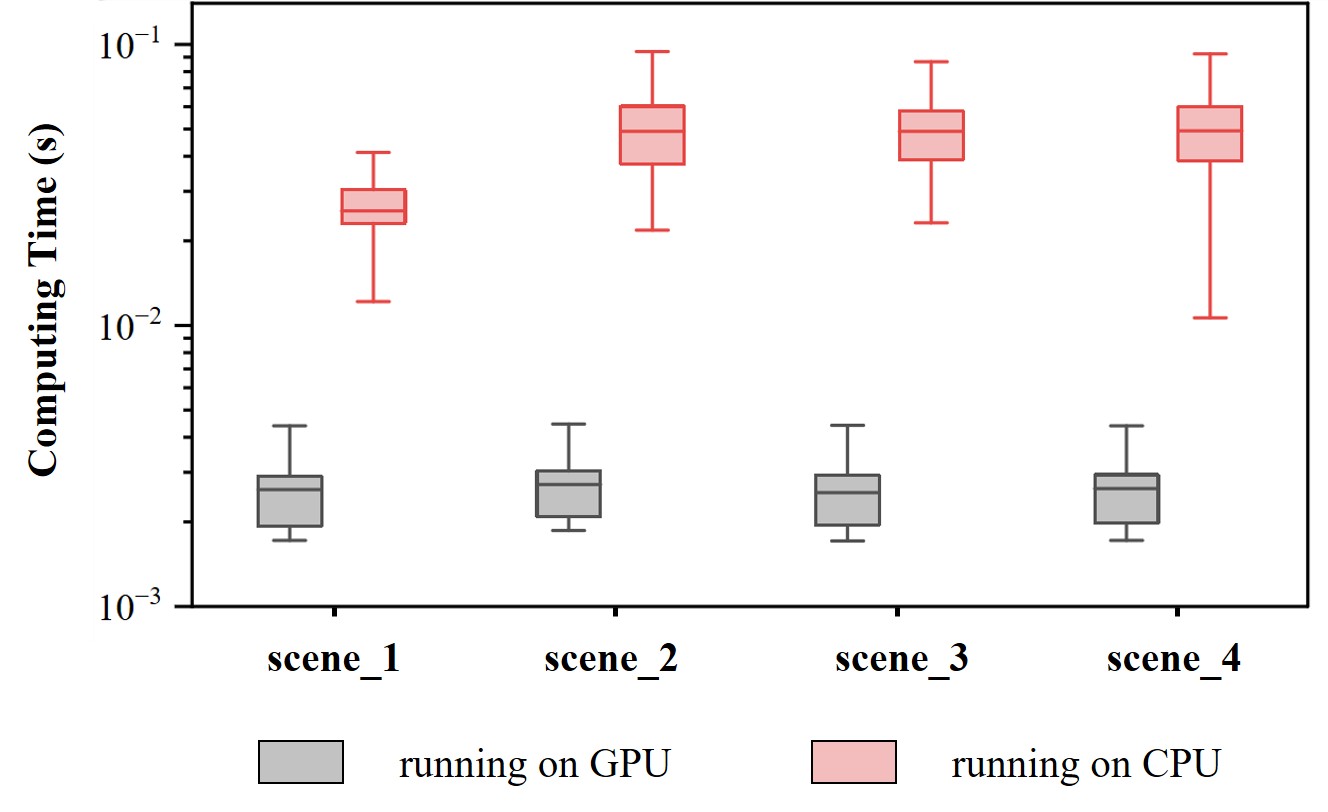}
		\caption{Box plots of the distribution of single execution times for local 3D grid map construction utilizing CPU and GPU across all four simulation scenarios.}
		\label{fig14}
	\end{center}
\end{figure}

As described in Sec.~\ref{Local 3D Grid Map Construction}, we leverage a highly parallelized GPU architecture to execute the construction process of the local 3D grid map, thereby effectively mitigating the massive computational overhead associated with dense 3D grid updates. Table~\ref{table5} and Fig.~\ref{fig14} present the average computational overhead and the distribution of per-run execution times for constructing the local 3D grid map utilizing both GPU and CPU across the four simulation scenarios. The experimental results demonstrate that constructing the local 3D grid map on a CPU requires approximately 18 times more computational time compared to the GPU implementation. This notable difference definitively validates the superiority of integrating a heterogeneous computing architecture within the proposed framework.

\subsection{Ablation Study of Viewpoint Generation Scheme}
\label{Ablation Study of Viewpoint Generation Scheme}

\begin{figure}[t]
	\begin{center}
		\includegraphics[width=\linewidth]{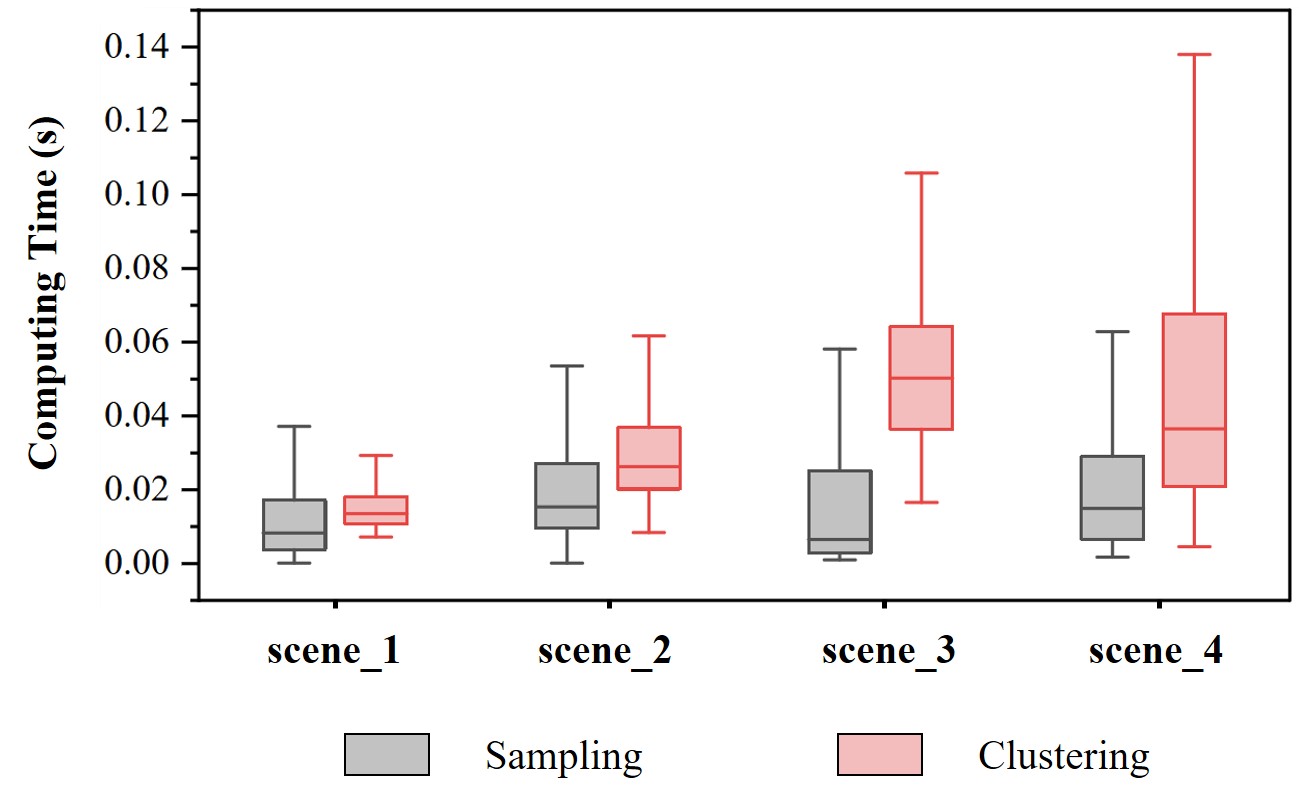}
		\caption{Box plots of the distribution of single execution times for the sampling and clustering viewpoint generation schemes across all four simulation scenarios.}
		\label{fig15}
	\end{center}
\end{figure}

As described in Sec. \ref{Viewpoint Generation}, uniformly sampling candidate points around the robot is an alternative viewpoint generation scheme, which is precisely the methodology adopted by FAEL \cite{huang2023fael}. However, this approach is highly computationally intensive as it necessitates sequentially attempting to associate every individual frontier with each sampled candidate viewpoint. This computational burden becomes particularly severe in sparsely occupied spaces, where the abundance of both candidate points and frontiers drastically escalates the total calculation time. Consequently, the proposed framework employs a strategy that first clusters the frontiers and subsequently determines a single optimal viewpoint for each frontier cluster. This structured procedure crucially aids in maintaining a relatively stable computational overhead. To validate this assertion, we replicated the sampling based viewpoint generation scheme utilized in FAEL to conduct a direct comparative analysis of computational overhead against our clustering strategy. When implementing the sampling methodology, we explicitly incorporated a comprehensive series of acceleration techniques adopted by FAEL, including the filtering of adjacent observation points, to ensure that the sampling based scheme operates with maximum possible computational efficiency. Table~\ref{table6} and Fig.~\ref{fig15} present the average computational overhead and the distribution of per run execution times for both viewpoint generation schemes across the four simulation scenarios. The experimental results demonstrate that within $scene\_3$ and $scene\_4$, which include sparsely occupied spaces, the computational expense of the sampling scheme is approximately twice that of the clustering method, thereby substantiating our earlier claims. Moreover, the computational overhead of the clustering scheme remains exceedingly consistent, whether deployed in corridor dominated layouts such as $scene\_2$ or expansive sparsely occupied spaces like $scene\_3$ and $scene\_4$. This remarkable consistency definitively proves the inherent advantage of the clustering approach regarding overall computational stability.

\begin{table}[t]
	\caption{Average Computing Time for Local 3D Grid Map Construction Using CPU and GPU Architectures}
	\label{table5}
	\centering
	\begin{tabular}{p{2.2cm}<{\centering}|p{1.0cm}<{\centering}p{1.0cm}<{\centering}p{1.0cm}<{\centering}p{1.0cm}<{\centering}}
		\toprule
		& \textbf{scene\_1} & \textbf{scene\_2} & \textbf{scene\_3} & \textbf{scene\_4} \\ \hline
		running on GPU & 0.0027\,s            & 0.0028\,s            & 0.0027\,s            & 0.0027\,s            \\
		running on CPU & 0.0268\,s            & 0.0506\,s            & 0.0500\,s            & 0.0504\,s            \\ \bottomrule
	\end{tabular}
\end{table}

\begin{table}[t]
	\caption{Average Computing Time of the Sampling and Clustering Viewpoint Generation Schemes}
	\label{table6}
	\centering
	\begin{tabular}{p{1.5cm}<{\centering}|p{1.0cm}<{\centering}p{1.0cm}<{\centering}p{1.0cm}<{\centering}p{1.0cm}<{\centering}}
		\toprule
		& \textbf{scene\_1} & \textbf{scene\_2} & \textbf{scene\_3} & \textbf{scene\_4} \\ \hline
		clustering & 0.0012\,s            & 0.0024\,s            & 0.0026\,s            & 0.0022\,s            \\
		sampling & 0.0015\,s            & 0.0029\,s            & 0.0053\,s            & 0.0051\,s            \\ \bottomrule
	\end{tabular}
\end{table}

\subsection{Ablation Study of Center Alignment for Staircases}
\label{Ablation Study of Center Alignment for Staircases}

\begin{figure}
	\begin{center}
		\includegraphics[width=\linewidth]{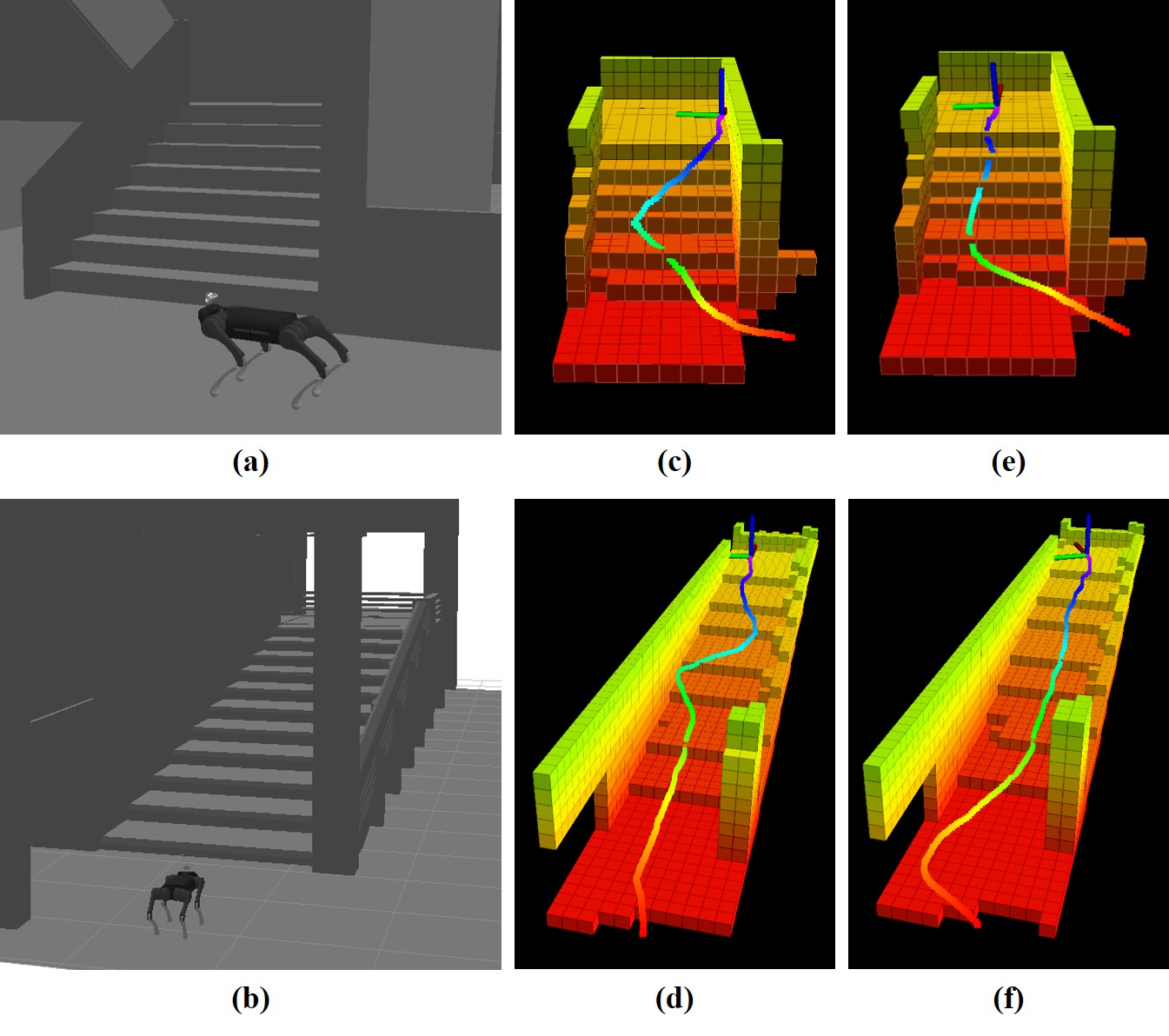}
		\caption{Comparative visualization of the quadruped robot ascending staircases. (a) and (b) illustrate the selected staircase regions for evaluation. (c) and (d) demonstrate the erratic climbing trajectories characterized by yaw angle variations generated without the center alignment constraint. (e) and (f) display the nearly straight trajectories produced when the proposed strategy is applied.}
		\label{fig16}
	\end{center}
\end{figure}

\begin{figure}
	\begin{center}
		\includegraphics[width=\linewidth]{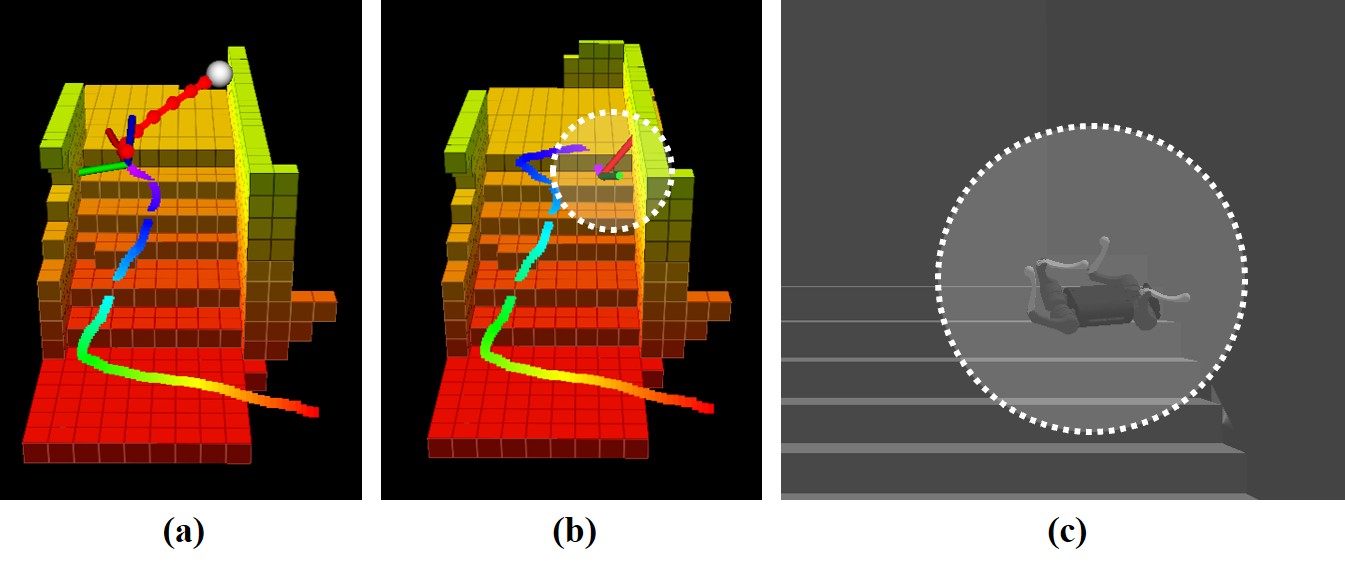}
		\caption{Sequential illustration of a lateral rollover failure caused by the absence of the center alignment constraint. (a) While the quadruped robot climbs facing left, a viewpoint distributed on the right is selected as the next navigation goal, compelling the robot to physically maneuver toward it. (b) and (c) To reach this new target, the robot is forced to substantially alter its yaw angle directly upon the staircase, a dangerous kinematic shift that destabilizes the platform and results in a severe fall.}
		\label{fig17}
	\end{center}
\end{figure}

As discussed in Sec.~\ref{Introduction}, the arbitrary spatial distribution of viewpoints within staircases can induce drastic yaw angle variations while the robot climbs stairs, thereby exposing the quadruped platform to a risk of falling. To mitigate this critical issue, we propose a center alignment strategy in Sec.~\ref{Center Alignment for Staircases}. This strategy constrains the distribution of viewpoints to the geometric center of each stair tread, preventing the quadruped robot from oscillating laterally and experiencing severe yaw angle variations during autonomous exploration across staircase regions. For experimental validation, we specifically select the first flight of the corner staircase in $scene\_2$ and the primary staircase connecting the first and second floors in $scene\_4$ as our targeted exploration regions, as depicted in Fig.~\ref{fig16} (a) and Fig.~\ref{fig16} (b). By executing autonomous exploration within these staircases, we aim to verify the trajectory improvements yielded by the proposed center alignment strategy. The staircases in $scene\_1$ and $scene\_3$ are omitted from this specific evaluation since their geometric dimensions and inclination angles are highly similar to those found in $scene\_4$. Fig.~\ref{fig16} (c) and Fig.~\ref{fig16} (d) demonstrate that in the absence of the center alignment strategy, the robot exhibits significant yaw angle variations during staircase climbing. Upon implementing the center alignment constraint, this kinematic instability is significantly alleviated, allowing the quadruped robot to traverse the stairs along a nearly straight trajectory (as illustrated in Fig.~\ref{fig16} (e) and Fig.~\ref{fig16} (f)). We evaluated each configuration across 20 independent trials in $scene\_2$ and $scene\_4$. Throughout all trials, the center alignment strategy consistently guaranteed a stable and nearly straight climbing trajectory. Conversely, operating without the center alignment constraint resulted in a catastrophic fall during one of the 20 trials in $scene\_2$. As illustrated in Fig.~\ref{fig17} (a), while the quadruped platform was climbing with a leftward orientation, a viewpoint distributed on the far right was selected as the next immediate navigation goal. This forced the robot to execute an aggressive yaw rotation directly upon the staircase to orient itself toward the new target. This highly perilous kinematic maneuver ultimately caused the quadruped robot to lose its balance and experience a lateral rollover, as captured in Fig.~\ref{fig17} (b) and Fig.~\ref{fig17} (c). Although this severe failure occurred only once out of 20 trials, the potential consequences of such a fall are strictly unacceptable, as it carries a substantial probability of permanently damaging the delicate LiDAR sensors and the onboard computing hardware. This critical failure case further underscores the absolute necessity of the proposed center alignment strategy specifically designed for staircases.

\subsection{Ablation Study of Dual Path Searching}
\label{Ablation Study of Dual Path Searching}

\begin{figure}
	\begin{center}
		\includegraphics[width=\linewidth]{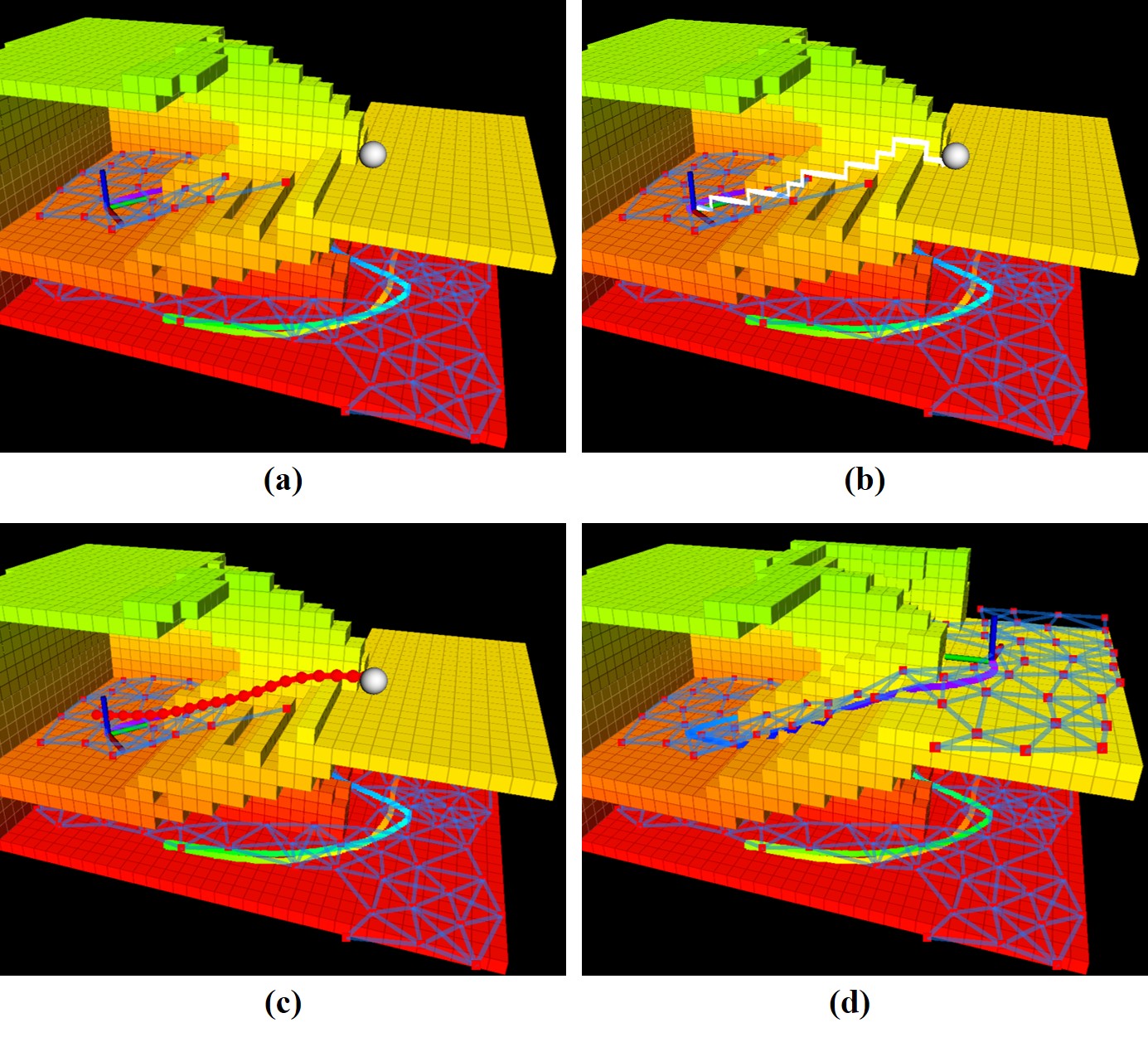}
		\caption{Sequential demonstration of the dual path searching mechanism solving a topological disconnection on a corner staircase. (a) A target viewpoint fails to link to the topological graph. (b) The traversable grid-based A-star search activates as a reliable fallback, successfully finding a supplementary path indicated by the solid white line. (c) The local planner generates a navigation trajectory based on this discovered route. (d) The topological graph is fully recovered and reconnected after the robot climbs and densely scans the staircase region in close proximity.}
		\label{fig18}
	\end{center}
\end{figure}

As discussed in Sec.~\ref{Dual Path Searching}, the dual path searching mechanism effectively prevents path search failures between the robot position at the bottom of the stairs and the target viewpoint at the top, which are typically caused by topological disconnections in the staircase region. We utilize the complete corner staircase in $scene\_2$ as our experimental environment to validate the efficacy of the dual path searching approach, since corner staircases represent the architectural structure most susceptible to topological breakages. When the robot finishes climbing the first flight and arrives at the half landing, the point cloud scanned toward the second flight lacks temporal accumulation and is therefore exceedingly sparse. Such sparse point clouds inherently yield degraded tomography maps, which subsequently result in poorly constructed topological graphs. As illustrated in Fig.~\ref{fig18} (a), the target viewpoint located above the stairs cannot be linked into the topological graph due to failed collision-free verification. At this juncture, the A-star search algorithm operating on the traversable grid is activated, successfully discovering a feasible path to the target point, as depicted by the solid white line in Fig.~\ref{fig18} (b). Based on this discovered path, the local planner generates the corresponding navigation trajectory, as shown in Fig.~\ref{fig18} (c). Once the quadruped robot successfully climbs the stairs following this trajectory, the topological graph within the staircase comprehensively recovers because the entire structural region is scanned by the LiDAR sensor at a close proximity, as illustrated in Fig.~\ref{fig18} (d). We similarly conducted 20 independent test trials to directly compare the proposed dual path searching approach against an exclusively topology based searching method. The dual path strategy successfully and exhaustively explored the entire corner staircase region in all 20 attempts, whereas the purely topological method succeeded merely 12 times. This quantitative evidence further proves that the proposed methodology significantly enhances the robustness of the exploration framework when operating within staircase environments.

\subsection{Real-World Experiment}
\label{Real-World Experiment}

\begin{figure}
	\begin{center}
		\includegraphics[width=\linewidth]{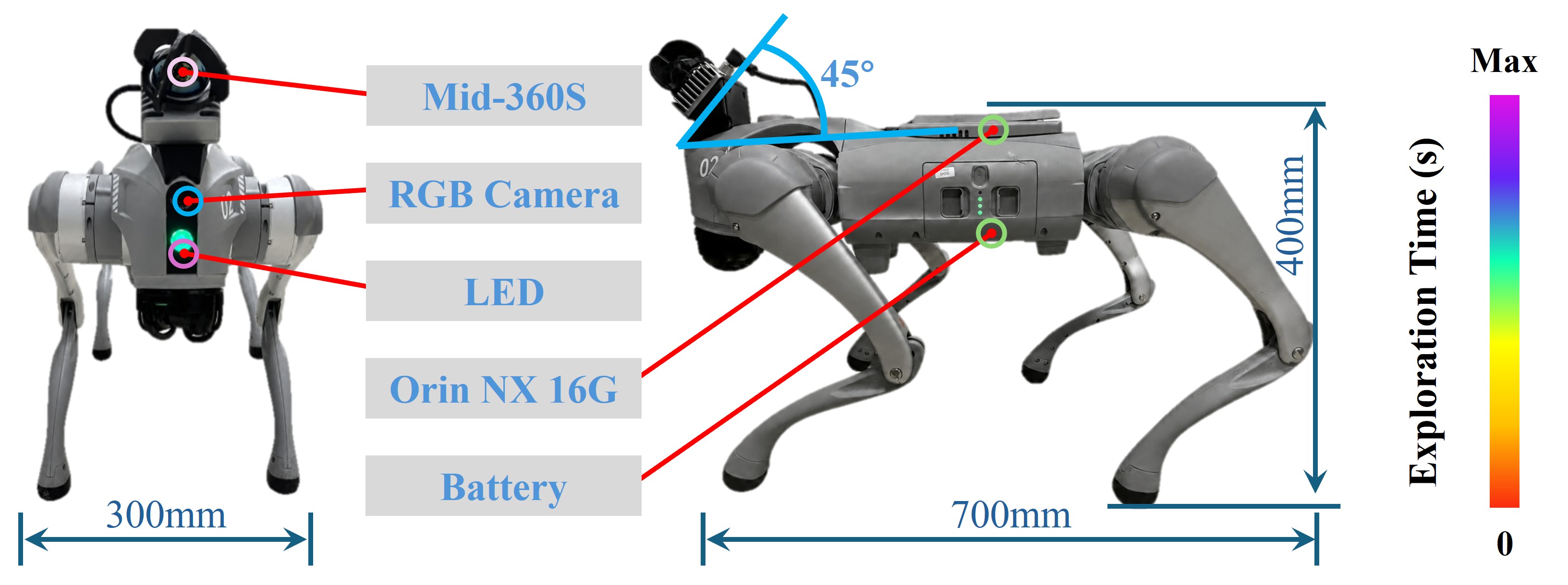}
		\caption{Hardware configuration of the robotic platform and the trajectory visualization scheme for the real-world experiments. The system comprises a Unitree Go2 quadruped robot equipped with a Mid-360S LiDAR and an onboard NVIDIA Jetson Orin NX computing unit. The accompanying rainbow color gradient applied to the trajectory intuitively illustrates the continuous chronological progression of the exploration process.}
		\label{fig19}
	\end{center}
\end{figure}

\begin{table}[]
	\caption{Configuration and Performance Metrics of the Four Real-World Testing Scenarios}
	\label{table7}
	\centering
	\begin{tabular}{c|cccc}
		\toprule
		& \textbf{Size (m)}         & \textbf{\begin{tabular}[c]{@{}c@{}}Exp.\\ Time (s)\end{tabular}} & \textbf{\begin{tabular}[c]{@{}c@{}}Traj.\\ Dist. (m)\end{tabular}} & \textbf{\begin{tabular}[c]{@{}c@{}}Max\\ Vel. (m/s)\end{tabular}} \\ \hline
		$stairwell$    & 5.5$\times$14$\times$21.5 & 233.9                                                            & 104.9                                                              & 1.0                                                               \\
		$building\_1$  & 16$\times$38$\times$8     & 290.0                                                            & 132.2                                                              & 0.75                                                              \\
		$building\_2$ & 51$\times$45$\times$8      & 375.0                                                            & 166.0                                                              & 0.75                                                              \\
		$lobby$        & 35$\times$26$\times$5     & 247.0                                                            & 99.9                                                               & 1.0                                                               \\ \bottomrule
	\end{tabular}
\end{table}

We further conduct experiments to validate the practical feasibility of RAEM for autonomous exploration in real-world environments. As illustrated in Fig.~\ref{fig19}, a Unitree Go2 is employed as the quadruped robot platform. A Mid-360S LiDAR is mounted on the head of the robot with a 45$^\circ$ downward pitch angle, and its effective sensing range is strictly configured to 4.0\,m to ensure consistency with the simulation experiments. All computations are executed directly on the standard onboard NVIDIA Jetson Orin NX 16GB processing unit. To provide the crucial state estimation and dense point cloud data required by RAEM, we run Fast-LIO2 \cite{xu2022fast} to achieve robust real time localization and mapping. In all subsequent experimental results, we map the continuous temporal progression of the robot position onto a rainbow colored trajectory. This color gradient visually encodes the elapsed time, thereby providing an intuitive understanding of the dynamic spatial exploration process from start to finish. We selected four representative testing scenarios to systematically evaluate our approach. Detailed configuration and performance metrics for these environments, including their geometric sizes, the total time required to achieve complete exploration, the cumulative traveling distance, and the maximum allowable velocity of the quadruped robot, are comprehensively summarized in Table \ref{table7}. The demonstration of the dynamic exploration process is included in the supplementary video.

\begin{figure}
	\begin{center}
		\includegraphics[width=\linewidth]{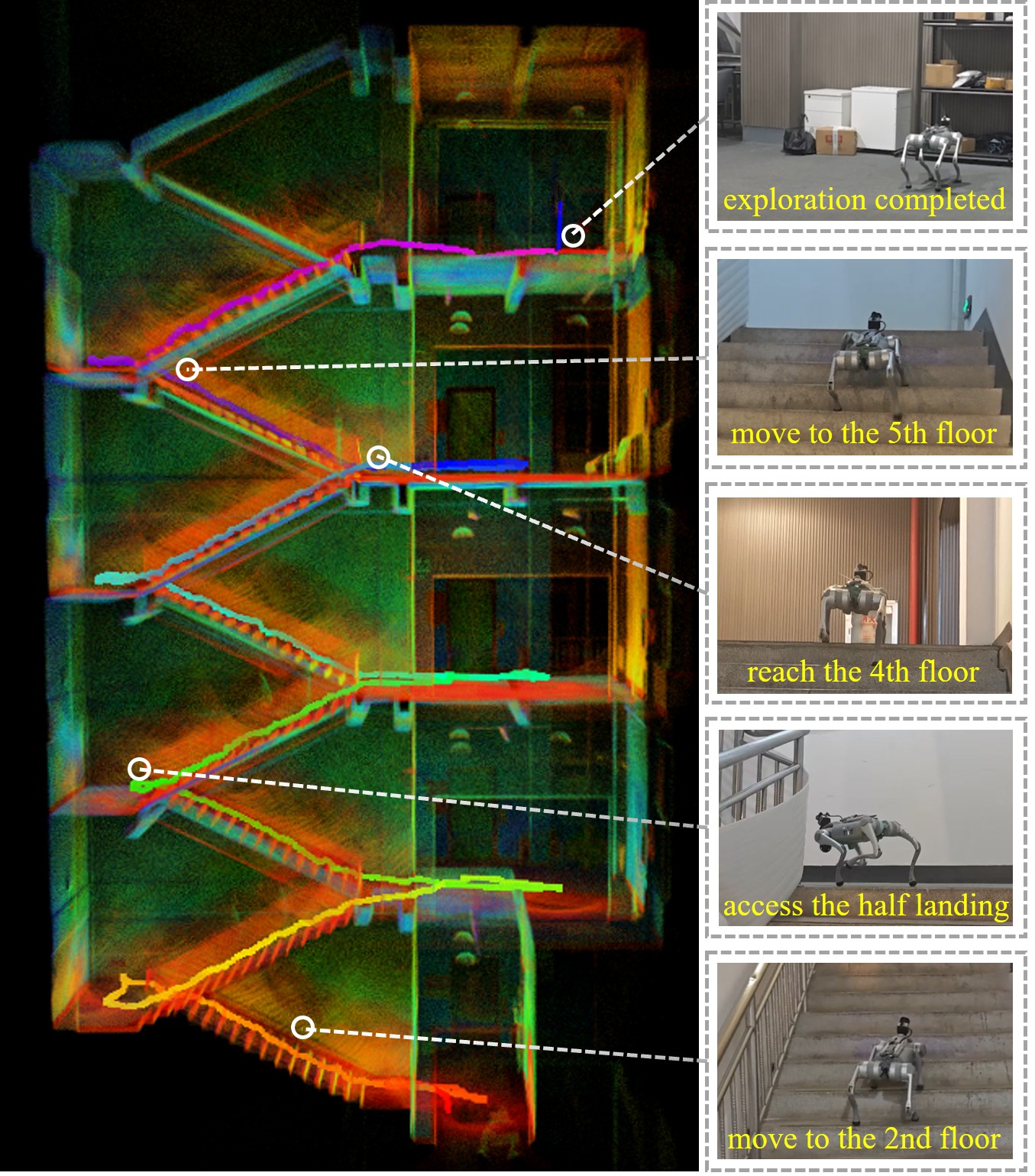}
		\caption{Experimental results in the real-world $stairwell$ scene. The proposed framework successfully enables the quadruped robot to autonomously explore from the first floor to the fifth floor.}
		\label{fig20}
	\end{center}
\end{figure}

The $stairwell$ scene strictly confines the exploration boundary within a building stairwell spanning from the first to the fifth floor, serving to evaluate the robustness of the proposed framework when addressing climbing challenges during exploration. The experimental results presented in Fig.~\ref{fig20} demonstrate that RAEM successfully enables the quadruped robot to autonomously explore from the first floor to the fifth floor, successfully climbing four consecutive corner staircases. This success firmly validates the exceptional robustness of the proposed system, as any viewpoint generation failure at intermediate landings or path searching failure would immediately halt the robot and terminate the exploration. Although the ablation studies conducted within the simulation environments in Sec. \ref{Ablation Study of Dual Path Searching} have previously demonstrated the robustness of RAEM when handling corner staircases, the field results obtained in the $stairwell$ scene explicitly prove that this algorithmic robustness is preserved even when deployed into real-world environments inherently characterized by map noise and localization errors.

\begin{figure}[t]
	\begin{center}
		\includegraphics[width=\linewidth]{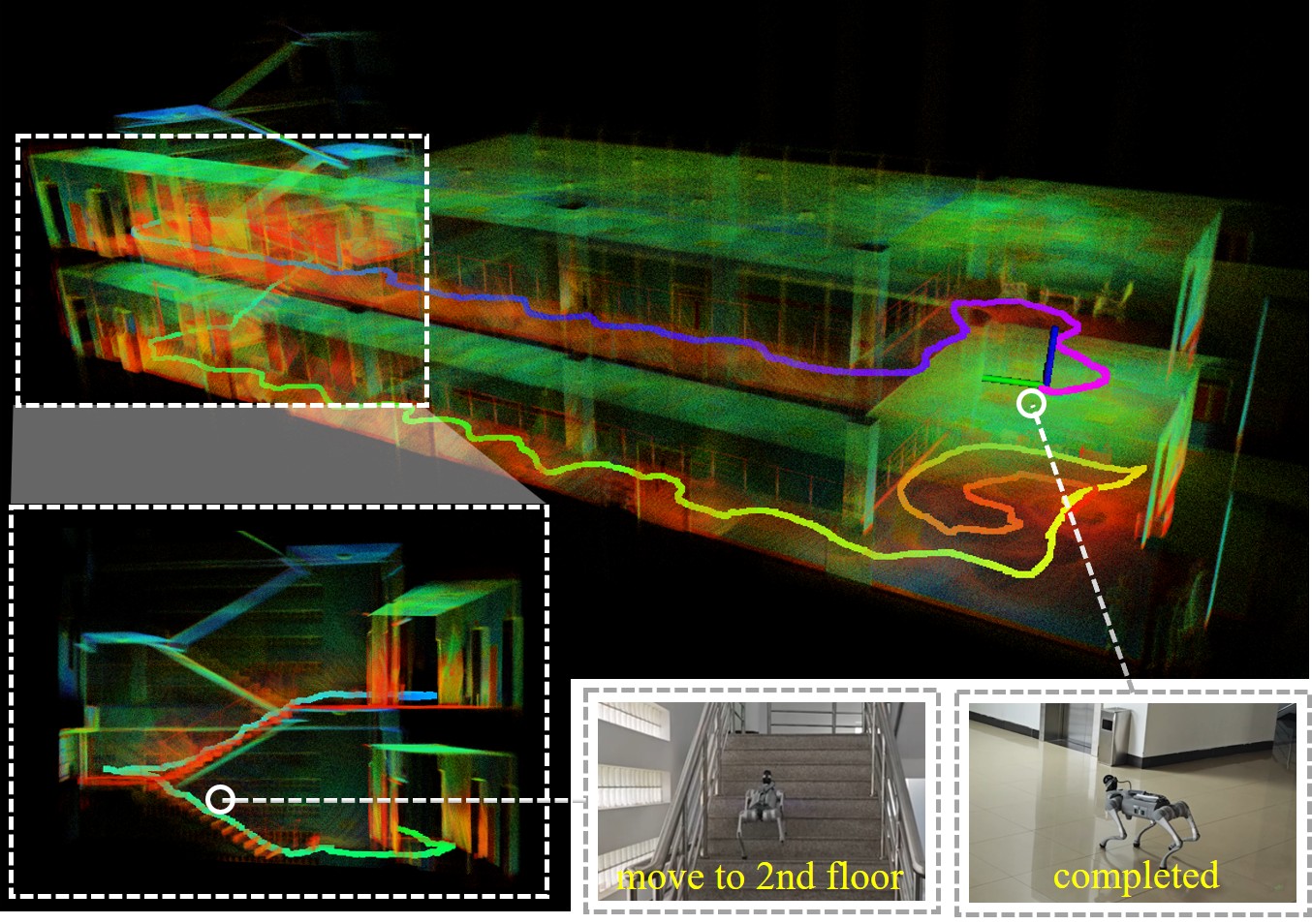}
		\caption{Experimental results in the real-world $building\_1$ scene. Upon discovering a staircase within the designated exploration boundary, the quadruped robot autonomously climbs to the upper floor to seamlessly continue exploration.}
		\label{fig21}
	\end{center}
\end{figure}

\begin{figure}[t]
	\begin{center}
		\includegraphics[width=\linewidth]{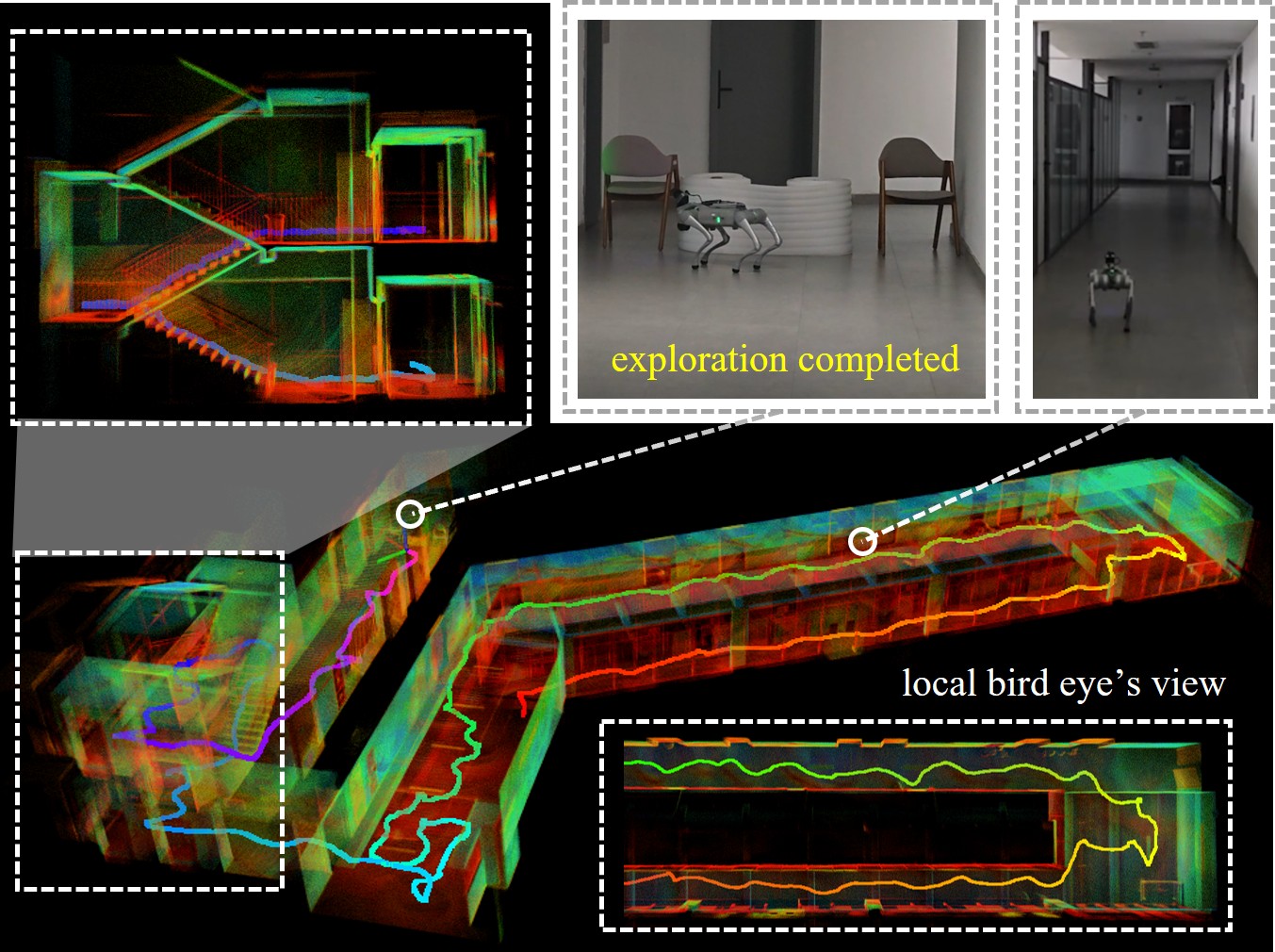}
		\caption{Experimental results in the real-world $building\_2$ scene.}
		\label{fig22}
	\end{center}
\end{figure}

The $building\_1$ and $building\_2$ scenes designate two adjacent floors within each building as targeted exploration areas to test the comprehensive performance of exploring multi-floor environments. To challenge the proposed framework, the quadruped robot was deliberately initialized at a location far from the staircases. This placement ensures that when the stair regions are first discovered, they lack sufficient LiDAR observation, intentionally forcing the framework to process highly sparse and fragmented point clouds for staircases. The experimental results presented in Fig.~\ref{fig21} and Fig.~\ref{fig22} demonstrate that despite these perceptual adversities, RAEM successfully executes the complete exploration across both floors in both $building\_1$ and $building\_2$. Specifically, during the $building\_2$ trial, a physical barricade was manually introduced near the conclusion of the task to obstruct access to a prolonged forward corridor. This deliberate intervention was necessary to prevent the severe state estimation degradation that Fast-LIO2 inevitably experiences when operating within such extended featureless geometries.

\begin{figure}[t]
	\begin{center}
		\includegraphics[width=\linewidth]{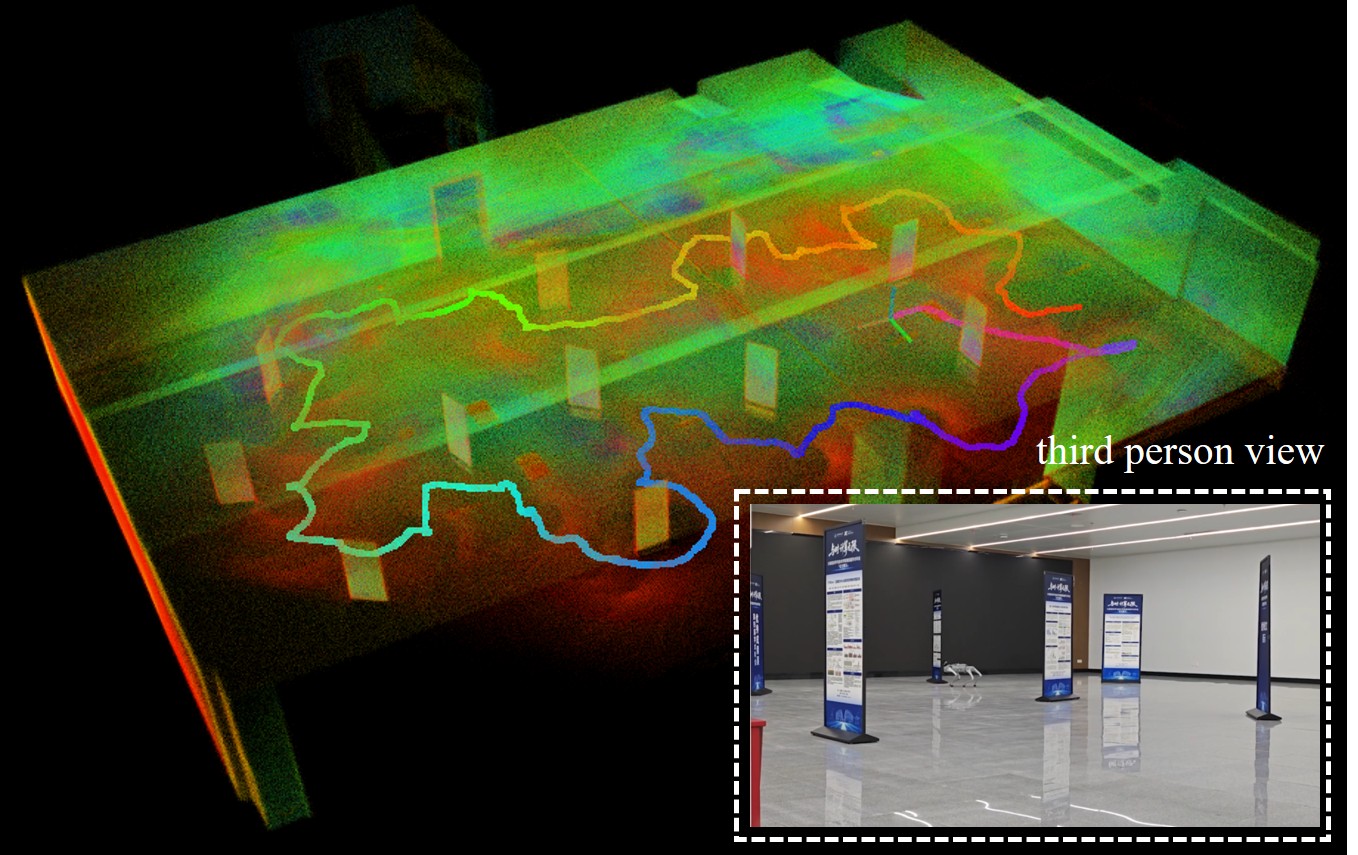}
		\caption{Experimental results in the real-world $lobby$ scene. Despite the sparsely occupied spaces generating numerous frontiers and candidate viewpoints that demand higher computational overhead, the quadruped robot maintains continuous movement, demonstrating the computational stability of the proposed framework.}
		\label{fig23}
	\end{center}
\end{figure}

The $lobby$ scene is specifically selected to evaluate the computational stability of the proposed framework when operating within sparsely occupied spaces. As thoroughly discussed in Sec.~\ref{Ablation Study of Viewpoint Generation Scheme}, such expansive open areas naturally produce an immense volume of frontiers and candidate viewpoints, which accordingly demands higher computational overhead. However, the experimental results presented in Fig.~\ref{fig23} demonstrate that the sparsely occupied scene does not affect the stability of RAEM. Throughout the entire process, the quadruped robot maintains continuous movement, with no pauses for waiting planning results. The supplementary video provides an intuitive and dynamic demonstration of this continuous exploration capability in sparsely occupied spaces.

\subsection{Limitation}
\label{Limitation}

Currently, the proposed framework primarily focuses on upward staircase exploration and does not explicitly address downward cross-floor traversal. This limitation mainly arises from both perception and locomotion considerations. Due to the fixed mounting pitch of the onboard LiDAR, downward staircases can remain within the sensor's perceptual blind spot. Sufficient observations of downward structures are typically available only when the robot approaches the upper edge of a staircase with an appropriate sensing direction, which cannot always be ensured during autonomous exploration. Furthermore, compared with ascending traversal, quadruped descending locomotion imposes higher requirements on dynamic stability and control accuracy, making reliable autonomous execution more challenging.

\section{Conclusion}
\label{Conclusion}

In this paper, we propose RAEM, a robust autonomous exploration framework tailored for quadruped robots operating in multi-floor environments. The core idea is a hybrid local-global traversability representation, in which a local tomography map and an explicitly categorized local 3D grid map are used for online terrain analysis and connectivity evaluation, while an elevation-aware global topological graph is incrementally constructed and updated to represent traversable connectivity across the explored environment. This design avoids maintaining a computationally expensive global tomography map while supporting efficient cross-floor exploration planning. To address perceptual degradation in stairwells, we further introduce a staircase center alignment strategy and a dual path searching mechanism. The former reduces abrupt yaw variations caused by irregularly distributed viewpoints, while the latter provides alternative guidance paths when topological connectivity is temporarily disrupted by sparse and fragmented observations. The framework is implemented using a heterogeneous CPU-GPU architecture to support efficient online computation. Extensive simulation and real-world experiments demonstrate that RAEM enables continuous autonomous exploration across multi-floor environments, including corner staircases and a five-floor stairwell. Future work includes extending the proposed framework into a multi-robot collaborative cross-floor autonomous exploration scheme to facilitate highly efficient spatial mapping across large scale architectural structures.


\bibliographystyle{IEEEtran}
\bibliography{IEEEabrv,IEEEExample}

\begin{thebibliography}{10}
\providecommand{\url}[1]{#1}
\csname url@samestyle\endcsname
\providecommand{\newblock}{\relax}
\providecommand{\bibinfo}[2]{#2}
\providecommand{\BIBentrySTDinterwordspacing}{\spaceskip=0pt\relax}
\providecommand{\BIBentryALTinterwordstretchfactor}{4}
\providecommand{\BIBentryALTinterwordspacing}{\spaceskip=\fontdimen2\font plus
\BIBentryALTinterwordstretchfactor\fontdimen3\font minus
  \fontdimen4\font\relax}
\providecommand{\BIBforeignlanguage}[2]{{%
\expandafter\ifx\csname l@#1\endcsname\relax
\typeout{** WARNING: IEEEtran.bst: No hyphenation pattern has been}%
\typeout{** loaded for the language `#1'. Using the pattern for}%
\typeout{** the default language instead.}%
\else
\language=\csname l@#1\endcsname
\fi
#2}}
\providecommand{\BIBdecl}{\relax}
\BIBdecl

\bibitem{yuan2022sr}
Z.~Yuan, F.~Lang, T.~Xu, and X.~Yang, ``Sr-lio: Lidar-inertial odometry with
  sweep reconstruction,'' in \emph{2024 IEEE/RSJ International Conference on
  Intelligent Robots and Systems (IROS)}.\hskip 1em plus 0.5em minus
  0.4em\relax IEEE, 2024, pp. 7862--7869.

\bibitem{yuan2024sr}
Z.~Yuan, J.~Deng, R.~Ming, F.~Lang, and X.~Yang, ``Sr-livo:
  Lidar-inertial-visual odometry and mapping with sweep reconstruction,''
  \emph{IEEE Robotics and Automation Letters}, 2024.

\bibitem{zheng2024fast}
C.~Zheng, W.~Xu, Z.~Zou, T.~Hua, C.~Yuan, D.~He, B.~Zhou, Z.~Liu, J.~Lin,
  F.~Zhu \emph{et~al.}, ``Fast-livo2: Fast, direct lidar--inertial--visual
  odometry,'' \emph{IEEE Transactions on Robotics}, vol.~41, pp. 326--346,
  2024.

\bibitem{czimmermann2021autonomous}
T.~Czimmermann, M.~Chiurazzi, M.~Milazzo, S.~Roccella, M.~Barbieri, P.~Dario,
  C.~M. Oddo, and G.~Ciuti, ``An autonomous robotic platform for manipulation
  and inspection of metallic surfaces in industry 4.0,'' \emph{IEEE
  Transactions on Automation Science and Engineering}, vol.~19, no.~3, pp.
  1691--1706, 2021.

\bibitem{yang2025closed}
H.~Yang, R.~Tao, S.~Zhang, L.~Chen, T.~Zhang, M.~Liu, H.~Wang, and Y.~Wu,
  ``Closed-loop safety inspection application via embodied robots,'' in
  \emph{2025 2nd International Symposium on AI and Cybersecurity
  (ISAICS)}.\hskip 1em plus 0.5em minus 0.4em\relax IEEE, 2025, pp. 1--5.

\bibitem{shukla2025systematic}
V.~Shukla, A.~Shukla, S.~P. SK, and S.~Shukla, ``A systematic survey: role of
  deep learning-based image anomaly detection in industrial inspection
  contexts,'' \emph{Frontiers in Robotics and AI}, vol.~12, p. 1554196, 2025.

\bibitem{wang2025development}
L.~Wang, D.~Teng, R.~Jin, X.~Guo, X.~Meng, G.~Liu, X.~Luo, C.~Sun, and Z.~Su,
  ``The development of prefabricated buildings and intelligent construction
  based on digital twins,'' \emph{Journal of Intelligent Construction}, vol.~3,
  no.~1, pp. 1--25, 2025.

\bibitem{nochta2026participation}
T.~Nochta and K.~Oti-Sarpong, ``Participation matters: The social construction
  of digital twins for cities,'' \emph{Environment and Planning B: Urban
  Analytics and City Science}, vol.~53, no.~4, pp. 765--777, 2026.

\bibitem{zhou2025multi}
J.~Zhou, ``Multi-dimensional model and interactive simulation of intelligent
  construction based on digital twins,'' \emph{Scientific Reports}, vol.~15,
  no.~1, p. 32189, 2025.

\bibitem{cao2021tare}
C.~Cao, H.~Zhu, H.~Choset, and J.~Zhang, ``Tare: A hierarchical framework for
  efficiently exploring complex 3d environments,'' in \emph{Robotics: Science
  and Systems}, vol.~5, 2021, p.~2.

\bibitem{huang2023fael}
J.~Huang, B.~Zhou, Z.~Fan, Y.~Zhu, Y.~Jie, L.~Li, and H.~Cheng, ``Fael: Fast
  autonomous exploration for large-scale environments with a mobile robot,''
  \emph{IEEE Robotics and Automation Letters}, vol.~8, no.~3, pp. 1667--1674,
  2023.

\bibitem{long2024hphs}
S.~Long, Y.~Li, C.~Wu, B.~Xu, and W.~Fan, ``Hphs: hierarchical planning based
  on hybrid frontier sampling for unknown environments exploration,'' in
  \emph{2024 IEEE/RSJ International Conference on Intelligent Robots and
  Systems (IROS)}.\hskip 1em plus 0.5em minus 0.4em\relax IEEE, 2024, pp.
  12\,056--12\,063.

\bibitem{patruno2021robust}
C.~Patruno, V.~Ren{\`o}, N.~Mosca, M.~di~Summa, and M.~Nitti, ``A robust method
  for 2d occupancy map building for indoor robot navigation,'' in
  \emph{Multimodal Sensing and Artificial Intelligence: Technologies and
  Applications II}, vol. 11785.\hskip 1em plus 0.5em minus 0.4em\relax SPIE,
  2021, pp. 56--67.

\bibitem{ren2026real}
M.~Ren, X.~Zhang, B.~Liu, and D.~Gu, ``Real-time monocular 2d occupancy grid
  mapping for autonomous navigation of ground robots,'' \emph{Journal of Field
  Robotics}, vol.~43, no.~3, pp. 1844--1860, 2026.

\bibitem{fankhauser2014robot}
P.~Fankhauser, M.~Bloesch, C.~Gehring, M.~Hutter, and R.~Siegwart,
  ``Robot-centric elevation mapping with uncertainty estimates,'' in
  \emph{Mobile Service Robotics}.\hskip 1em plus 0.5em minus 0.4em\relax World
  Scientific, 2014, pp. 433--440.

\bibitem{fankhauser2018probabilistic}
P.~Fankhauser, M.~Bloesch, and M.~Hutter, ``Probabilistic terrain mapping for
  mobile robots with uncertain localization,'' \emph{IEEE Robotics and
  Automation Letters}, vol.~3, no.~4, pp. 3019--3026, 2018.

\bibitem{frey2022locomotion}
J.~Frey, D.~Hoeller, S.~Khattak, and M.~Hutter, ``Locomotion policy guided
  traversability learning using volumetric representations of complex
  environments,'' in \emph{2022 IEEE/RSJ International Conference on
  Intelligent Robots and Systems (IROS)}.\hskip 1em plus 0.5em minus
  0.4em\relax IEEE, 2022, pp. 5722--5729.

\bibitem{wang2023towards}
J.~Wang, L.~Xu, H.~Fu, Z.~Meng, C.~Xu, Y.~Cao, X.~Lyu, and F.~Gao, ``Towards
  efficient trajectory generation for ground robots beyond 2d environment,''
  \emph{arXiv preprint arXiv:2302.03323}, 2023.

\bibitem{yang2024efficient}
B.~Yang, J.~Cheng, B.~Xue, J.~Jiao, and M.~Liu, ``Efficient global navigational
  planning in 3-d structures based on point cloud tomography,'' \emph{IEEE/ASME
  Transactions on Mechatronics}, vol.~30, no.~1, pp. 321--332, 2024.

\bibitem{li2025real}
Y.~Li, K.~Chen, Y.~Wang, W.~Zhang, J.~Wang, H.~Chen, and Y.~Liu, ``Real-time
  multilevel terrain-aware path planning for ground mobile robots in
  large-scale rough terrains,'' \emph{IEEE Transactions on Robotics}, vol.~41,
  pp. 4159--4179, 2025.

\bibitem{lee2025trg}
D.~Lee, I.~M.~A. Nahrendra, M.~Oh, B.~Yu, and H.~Myung, ``Trg-planner:
  Traversal risk graph-based path planning in unstructured environments for
  safe and efficient navigation,'' \emph{IEEE Robotics and Automation Letters},
  vol.~10, no.~2, pp. 1736--1743, 2025.

\bibitem{zhou2021fuel}
B.~Zhou, Y.~Zhang, X.~Chen, and S.~Shen, ``Fuel: Fast uav exploration using
  incremental frontier structure and hierarchical planning,'' \emph{IEEE
  Robotics and Automation Letters}, vol.~6, no.~2, pp. 779--786, 2021.

\bibitem{yang2021graph}
F.~Yang, D.-H. Lee, J.~Keller, and S.~Scherer, ``Graph-based topological
  exploration planning in large-scale 3d environments,'' in \emph{2021 IEEE
  international conference on robotics and automation (ICRA)}.\hskip 1em plus
  0.5em minus 0.4em\relax IEEE, 2021, pp. 12\,730--12\,736.

\bibitem{tang2023bubble}
B.~Tang, Y.~Ren, F.~Zhu, R.~He, S.~Liang, F.~Kong, and F.~Zhang, ``Bubble
  explorer: Fast uav exploration in large-scale and cluttered 3d-environments
  using occlusion-free spheres,'' in \emph{2023 IEEE/RSJ International
  Conference on Intelligent Robots and Systems (IROS)}.\hskip 1em plus 0.5em
  minus 0.4em\relax IEEE, 2023, pp. 1118--1125.

\bibitem{duberg2020ufomap}
D.~Duberg and P.~Jensfelt, ``Ufomap: An efficient probabilistic 3d mapping
  framework that embraces the unknown,'' \emph{IEEE Robotics and Automation
  Letters}, vol.~5, no.~4, pp. 6411--6418, 2020.

\bibitem{musil2022spheremap}
T.~Musil, M.~Petrl{\'\i}k, and M.~Saska, ``Spheremap: Dynamic multi-layer graph
  structure for rapid safety-aware uav planning,'' \emph{IEEE Robotics and
  Automation Letters}, vol.~7, no.~4, pp. 11\,007--11\,014, 2022.

\bibitem{yu2023echo}
J.~Yu, H.~Shen, J.~Xu, and T.~Zhang, ``Echo: An efficient heuristic viewpoint
  determination method on frontier-based autonomous exploration for
  quadrotors,'' \emph{IEEE Robotics and Automation Letters}, vol.~8, no.~8, pp.
  5047--5054, 2023.

\bibitem{hornung2013octomap}
A.~Hornung, K.~M. Wurm, M.~Bennewitz, C.~Stachniss, and W.~Burgard, ``Octomap:
  An efficient probabilistic 3d mapping framework based on octrees,''
  \emph{Autonomous robots}, vol.~34, no.~3, pp. 189--206, 2013.

\bibitem{zhang2024falcon}
Y.~Zhang, X.~Chen, C.~Feng, B.~Zhou, and S.~Shen, ``Falcon: Fast autonomous
  aerial exploration using coverage path guidance,'' \emph{IEEE Transactions on
  Robotics}, vol.~41, pp. 1365--1385, 2024.

\bibitem{zhang2025fsmp}
S.~Zhang, X.~Zhang, Q.~Dong, Z.~Wang, H.~Xi, and J.~Yuan, ``Fsmp: A
  frontier-sampling-mixed planner for fast autonomous exploration of complex
  and large 3-d environments,'' \emph{IEEE Transactions on Instrumentation and
  Measurement}, 2025.

\bibitem{geng2025epic}
S.~Geng, Z.~Ning, F.~Zhang, and B.~Zhou, ``Epic: A lightweight lidar-based aav
  exploration framework for large-scale scenarios,'' \emph{IEEE Robotics and
  Automation Letters}, vol.~10, no.~5, pp. 5090--5097, 2025.

\bibitem{yuan2026aerial}
Z.~Yuan, Y.~Ren, Y.~Wang, L.~Zhao, S.~Sun, C.~Chen, L.~Zhu, X.~Yang, and K.-T.
  Cheng, ``Aerial exploration on point cloud maps via coverage path guidance,''
  \emph{IEEE Robotics and Automation Letters}, 2026.

\bibitem{kan2020online}
X.~Kan, H.~Teng, and K.~Karydis, ``Online exploration and coverage planning in
  unknown obstacle-cluttered environments,'' \emph{IEEE Robotics and Automation
  Letters}, vol.~5, no.~4, pp. 5969--5976, 2020.

\bibitem{wang2025tips}
Z.~Wang, S.~Pan, J.~Xu, X.~Tao, W.~Gao, and Q.~Wang, ``Tips: Tiered
  information-rich planning strategy for efficient ugv autonomous
  exploration,'' \emph{IEEE Robotics and Automation Letters}, 2025.

\bibitem{miki2022elevation}
T.~Miki, L.~Wellhausen, R.~Grandia, F.~Jenelten, T.~Homberger, and M.~Hutter,
  ``Elevation mapping for locomotion and navigation using gpu,'' in \emph{2022
  IEEE/RSJ International Conference on Intelligent Robots and Systems
  (IROS)}.\hskip 1em plus 0.5em minus 0.4em\relax IEEE, 2022, pp. 2273--2280.

\bibitem{meng2023terrainnet}
X.~Meng, N.~Hatch, A.~Lambert, A.~Li, N.~Wagener, M.~Schmittle, J.~Lee,
  W.~Yuan, Z.~Chen, S.~Deng \emph{et~al.}, ``Terrainnet: Visual modeling of
  complex terrain for high-speed, off-road navigation,'' \emph{arXiv preprint
  arXiv:2303.15771}, 2023.

\bibitem{triebel2006multi}
R.~Triebel, P.~Pfaff, and W.~Burgard, ``Multi-level surface maps for outdoor
  terrain mapping and loop closing,'' in \emph{2006 IEEE/RSJ international
  conference on intelligent robots and systems}.\hskip 1em plus 0.5em minus
  0.4em\relax IEEE, 2006, pp. 2276--2282.

\bibitem{applegate2011traveling}
D.~L. Applegate, R.~E. Bixby, V.~Chv{\'a}tal, and W.~J. Cook, ``The traveling
  salesman problem: a computational study,'' in \emph{The traveling salesman
  problem}.\hskip 1em plus 0.5em minus 0.4em\relax Princeton university press,
  2011.

\bibitem{croes1958method}
G.~A. Croes, ``A method for solving traveling-salesman problems,''
  \emph{Operations research}, vol.~6, no.~6, pp. 791--812, 1958.

\bibitem{zhou2020ego}
X.~Zhou, Z.~Wang, H.~Ye, C.~Xu, and F.~Gao, ``Ego-planner: An esdf-free
  gradient-based local planner for quadrotors,'' \emph{IEEE Robotics and
  Automation Letters}, vol.~6, no.~2, pp. 478--485, 2020.

\bibitem{xu2022fast}
W.~Xu, Y.~Cai, D.~He, J.~Lin, and F.~Zhang, ``Fast-lio2: Fast direct
  lidar-inertial odometry,'' \emph{IEEE Transactions on Robotics}, vol.~38,
  no.~4, pp. 2053--2073, 2022.

\end{thebibliography}

%
%
%
%
%
%
%
%
%

\end{document}